\documentclass[11pt]{article}
\pdfoutput=1

\usepackage{mathrsfs}
\usepackage{amsmath,amssymb}
\usepackage{bm}
\usepackage{natbib}
\usepackage[usenames]{color}
\usepackage{amsthm}

\usepackage{multirow} 
\usepackage{enumitem}

\usepackage{enumitem}
\usepackage{dsfont}
\usepackage{mathtools}

\usepackage{graphicx}  
\usepackage{float}  
\usepackage{subfigure}  

\usepackage[colorlinks,
linkcolor=red,
anchorcolor=blue,
citecolor=blue
]{hyperref}

\usepackage{mylatexstyle}

\usepackage{setspace}
\usepackage[left=1in, right=1in, top=1in, bottom=1in]{geometry}

\setlist[itemize]{itemsep=0pt, topsep=2pt}
\setlist[enumerate]{itemsep=-2pt, topsep=2pt}

\usepackage{xcolor}

\ifdefined\final
\usepackage[disable]{todonotes}
\else
\usepackage[textsize=tiny]{todonotes}
\fi
\allowdisplaybreaks

\title{Learning From Biased Soft Labels}

\author{Hua Yuan, Ning Xu, Yu Shi, Xin Geng and Yong Rui}

\date{}
\begin{document}
\maketitle
\begin{abstract}
    Knowledge distillation has been widely adopted in a variety of tasks and has achieved remarkable successes.
    Since its inception, many researchers have been intrigued by the dark knowledge hidden in the outputs of the teacher model.
    Recently, a study has demonstrated that knowledge distillation and label smoothing can be unified as learning from soft labels.
    Consequently, how to measure the effectiveness of the soft labels becomes an important question.
    Most existing theories have stringent constraints on the teacher model or data distribution, and many assumptions imply that the soft labels are close to the ground-truth labels.
    This paper studies whether biased soft labels are still effective.
    We present two more comprehensive indicators to measure the effectiveness of such soft labels.
    Based on the two indicators, we give sufficient conditions to ensure biased soft label based learners are \textit{classifier-consistent} and ERM \textit{learnable}.
   The theory is applied to three weakly-supervised frameworks.
    Experimental results validate that biased soft labels can also teach good students, which corroborates the soundness of the theory.
\end{abstract}

\section{Introduction}
\label{introduction}
Recently, knowledge distillation \cite{compression, ba2014deep, kd1} has engendered remarkable achievements in a wide range of applications.
Although it was firstly proposed for model compression by distilling knowledge from the big model (teacher) to the small model (student),
considerable efforts have been devoted to figuring out the dark knowledge hidden in the outputs of the teacher model.
The dark knowledge is compatibly utilized for transfer learning \cite{transfer1,transfer2,transfer3}.

In practice, the student loss is defined as the tradeoff between imitating the ground-truth label and imitating the output of the teacher model.
Many studies \cite{kdbetter1, kdbetter2} have demonstrated that learning from the teacher model can be more effective than the ground-truth labels.
This seems counterintuitive since it challenges the correctness of the ground-truth labels.
Apart from knowledge distillation, label smoothing \cite{ls_1, lsbetter1} also softens the labels by incorporating uniform noise, which is a useful trick to improve generalization.
Knowledge distillation and label smoothing are often analyzed together, and \citet{defective} elucidates that they can be unified as learning from soft labels.
The essence of both is why the soft labels are effective.

In this paper, we mainly focus on the effectiveness of these soft labels.
To be clarified, this paper investigates when the soft labels are effective, rather than when the soft labels are superior to ground-truth labels.
It is apparent that, when the soft labels are close to the ground-truth labels, the student model will have an adequate performance.
A straightforward question is, 

\begin{center}
   \boxed{``\textit{whether the large-bias soft labels are still effective?}".}
\end{center}

\begin{figure*}[t]
   \centering
   \hspace{-0.35in}
   \subfigure[]{
   \begin{minipage}[t]{0.25\linewidth}
   \centering
   \includegraphics[width=1.20in]{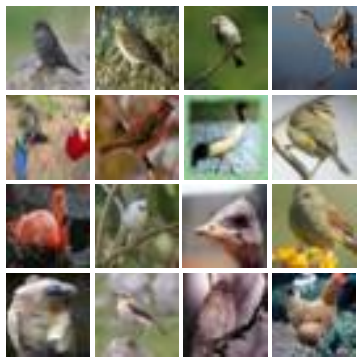}
   \end{minipage}%
   }%
   \subfigure[]{
   \begin{minipage}[t]{0.25\linewidth}
   \centering
   \includegraphics[width=1.70in]{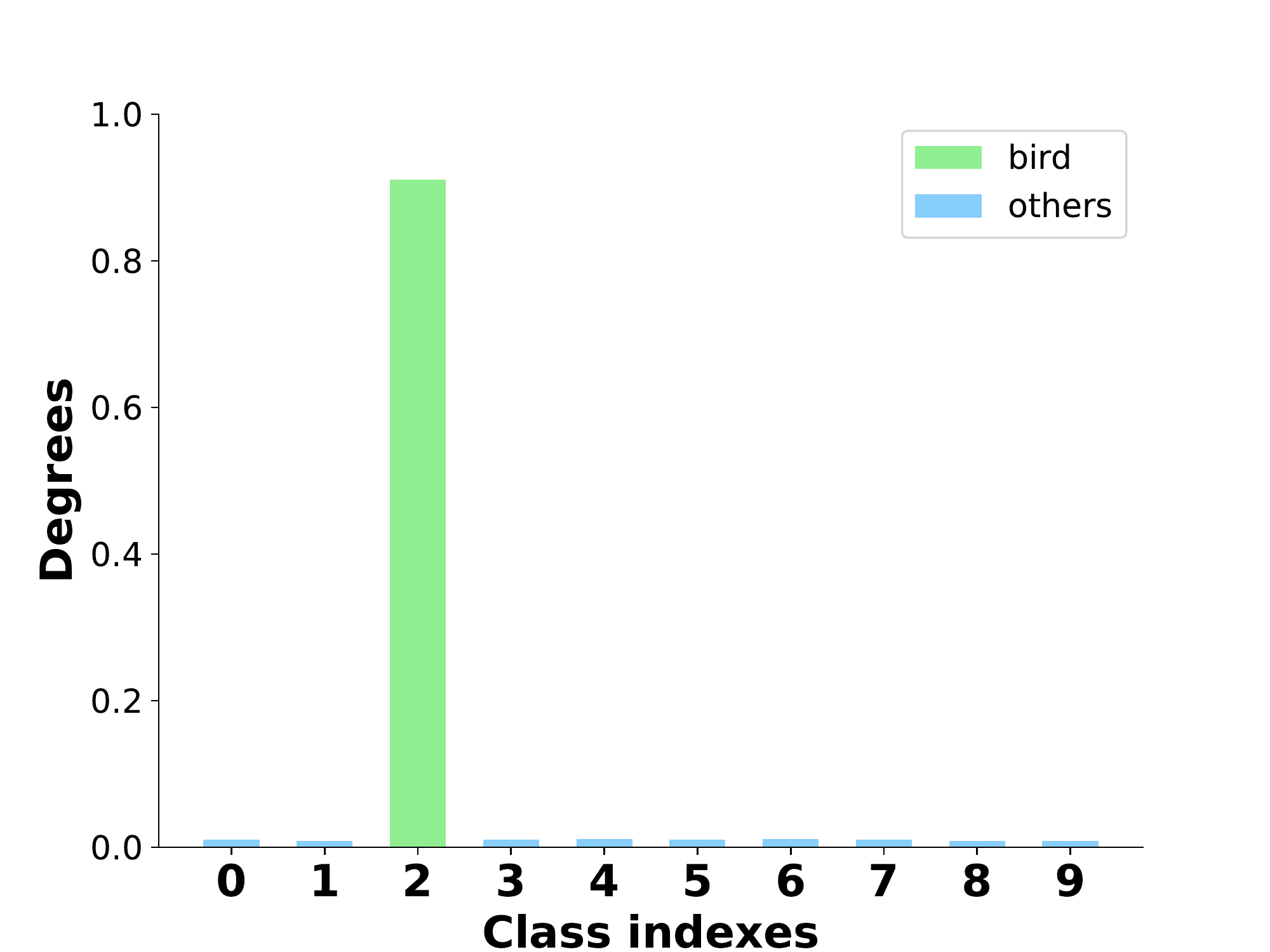}
   \end{minipage}%
   }%
   \subfigure[]{
   \begin{minipage}[t]{0.25\linewidth}
   \centering
   \includegraphics[width=1.70in]{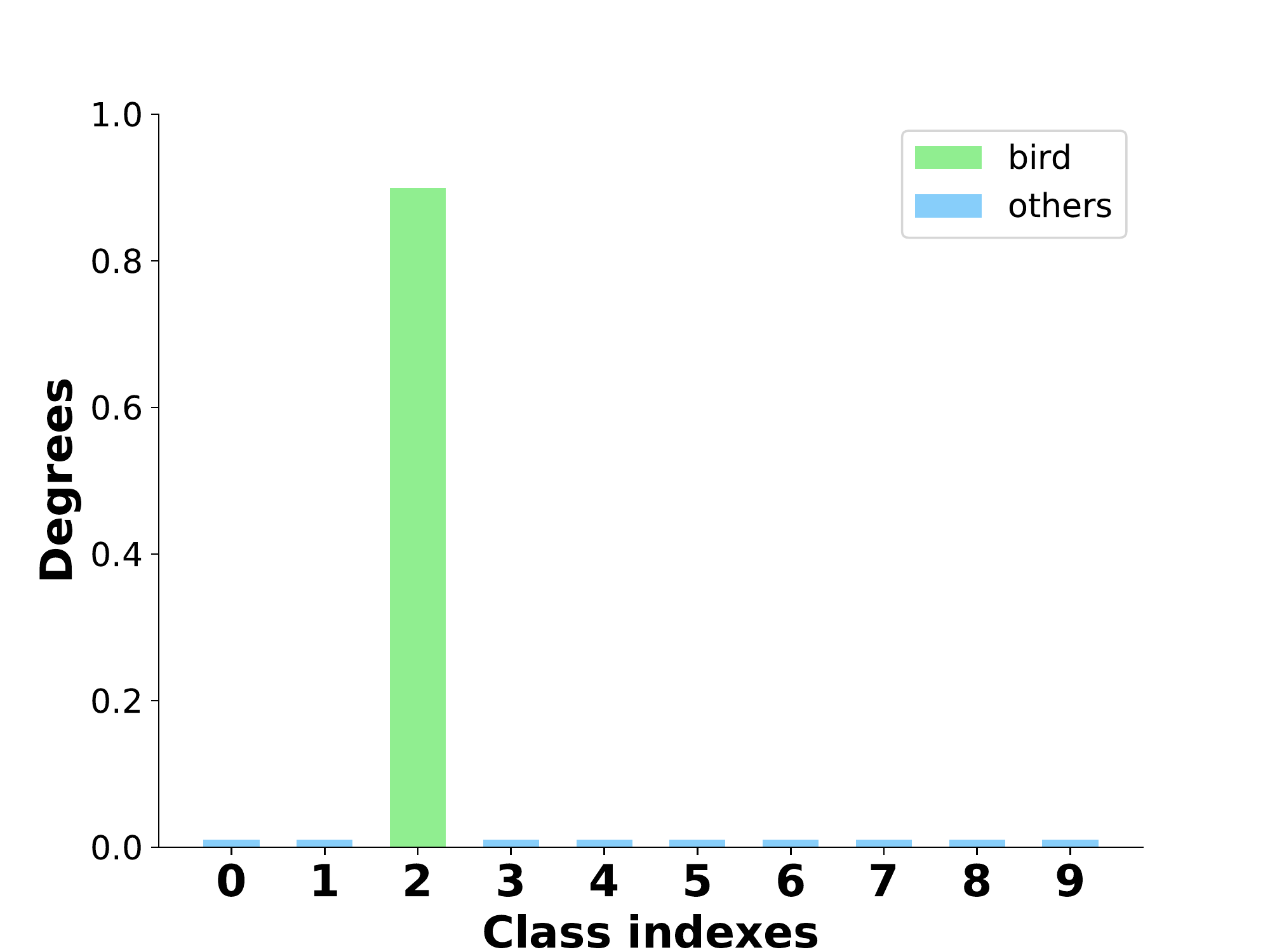}
   \end{minipage}
   }%
   \subfigure[]{
   \begin{minipage}[t]{0.25\linewidth}
   \centering
   \includegraphics[width=1.70in]{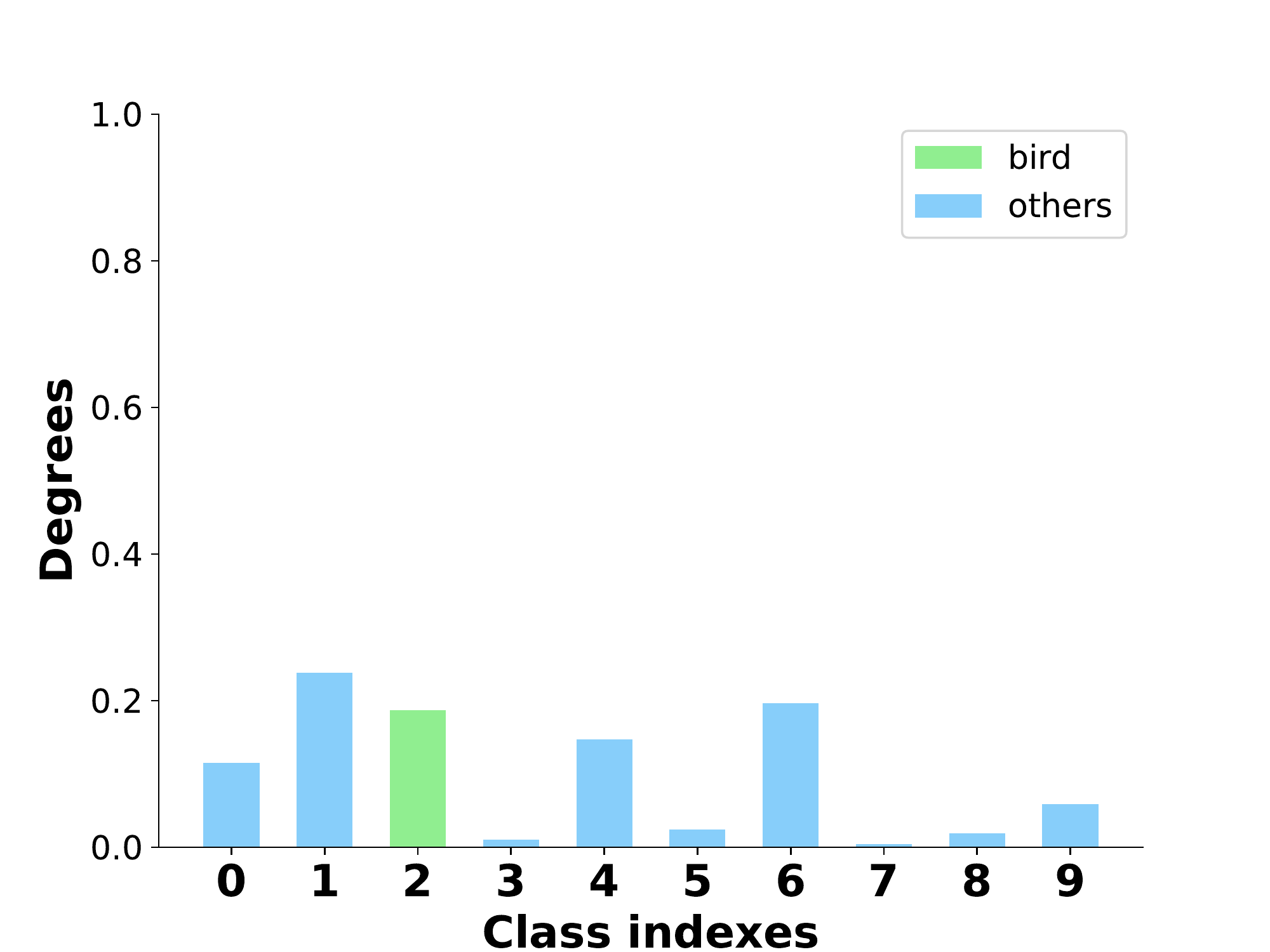}
   \end{minipage}%
   }%
   \centering
   \caption{(a) Images of the birds in CIFAR-10. (b) Defective knowledge distillation with $\alpha=0.9$ and $\tau=20$. (c) Label smoothing with $\alpha=0.9$. (d) Our customized soft labels.}
   \label{illustrate}
\end{figure*}

\citet{defective} empirically demonstrates the poorly-trained teacher model can also improve the student model.
However, it sets the tradeoff $\alpha=0.9$ and temperature $\tau=20$, which means the defective soft labels are still close to the ground-truth label.
Figure \ref{illustrate} illustrates defective soft labels, label smoothing (with $\alpha=0.9$) and our customized soft labels (detailed in subsection \ref{CSL}).

To measure the effectiveness of the soft labels, without accuracy, we propose two intuitive indicators, unreliability degree and ambiguity degree.
Furthermore, based on the two indicators, we prove that learning from the biased soft labels is \textit{classifier-consistent} and Empirical Risk Minimizing (ERM) \textit{learnable} under a moderate condition.
The theory is applicable not only to learning from poor teachers, but to all soft label based learners.
This result significantly extends the application scope of soft labels.
We apply it to three classic weakly-supervised frameworks: parital label learning, learning with additive noise, learning with incomplete data.

Among the weakly-supervised frameworks, soft labels of parital label learning and additive noise are spoiled by human or during collection.
In incomplete supervision, where only part of the data is labeled, the typical strategy is to label the unlabeled data and learn with all data iteratively.
Soft labels in these weakly-supervised frameworks are usually biased and we provide a guarantee for the learners in these fields.
Specifically, for the incomplete data, we delineate the dynamics of the model performance with an ideal accuracy funtion and give conditions to ensure the existence of the final accuracy.


To illustrate the soundness of our theory, we train the teacher models with some heuristic losses to generate soft labels with low accuracy but fulfilling the criteria in Theorem \ref{main}.
Training on these biased soft labels, the student model can achieve an adequate performance as if training on ground-truth labels, which is consistent with our theory.
In addition, the experiments of weakly-supervised learning also validate the effectiveness of biased soft labels.
Our contributions can be summarized as follows:
\begin{itemize}
   \item[$\bullet$] We focus on the effectiveness of soft labels and find that learning from biased soft labels may also achieve an adequate performance. A heuristic method is devised to generate biased soft labels that can train a good student.
   \item[$\bullet$] We give sufficient conditions to guarantee the effectiveness of the soft labels. It is proved that learning from such soft labels is classifier-consistent and ERM learnable. Experimental results validate our theory.
   \item[$\bullet$] Our theory is applied to three weakly-supervised frameworks where the soft labels are biased. In incomplete supervision, We delineate the dynamics of the model performance with an ideal accuracy function, and give the final accuracy.
\end{itemize}

\section{Related Work}
\label{rw}
\textbf{Knowledge Distillation and Label Smoothing} 
Knowledge Distillation (KD) has been widely adopted and achieved remarkable success since its inception.
It was firstly proposed in model compression and then applied to transfer learning.
There is growing interest in why distilling can transfer information and what is the \textit{dark knowledge} hidden in the soft labels.
Furthermore, the formalization of the dark knowledge is not restricted to vanilla knowledge distillation.
Self-distillation \cite{self1, self2} treats the mixture of the outputs and the ground-truth labels as targets, and the proportion is iteratively adjusted during training.
Ensemble KD \cite{ensemble1,ensemble2} employs the ensemble of the soft labels to improve generalization.
Besides, in mutual learning \cite{mutual,mutual2}, there is no explicit teacher network and 
multiple student learns from each other by synthesizing other soft labels.

Label Smoothing (LS) \cite{ls_1} is a regularization method to improve performance, where the soft labels are supposed to restrain overconfidence of the student model.
The essence behind KD and LS lies in the soft label, which are imitated by the student model.
Existing theories of the soft labels are diverse but there is no credible conclusion yet.
Many researches regard KD and LS as a regularization method to boost performance.
\citet{tang2020understanding} and \citet{LShelp} believe that the soft labels contain category knowledge which can help the student understand.
Besides, \citet{rethinking} assumes the ground-truth labels are sampled from Bayes prior probabilities and analyze the soft labels from a perspective of bias-variance tradeoff.
On the other hand, \citet{semiparametric} argued that the soft labels are best when they coincide with the Bayes probability distribution.
There are also papers \cite{noisydistillation,LShelp} showing that soft labels can mitigate noise.


\textbf{Label Enhancement}
Label Distribution Learning (LDL) \cite{ldl} was proposed to exploit the label distribution to mirror the relationship between the label and the instance, where the formalization of the label distribution is identical with the soft labels mentioned above.
Due to the high cost of quantifying the soft labels, Label Enhancement (LE) \cite{le} was proposed to recover the label distribution from the logical label by exploiting the implicit correlation among different labels.
In the following of the paper, we use nomenclature soft labels.
Numerous novel algorithms have been designed in recent years that aim to improve the predictive model with the soft labels \cite{asdfas, flefmll}.
\citet{LDLM} applied the margin theory to the soft labels and designed the adaptive margin loss.

In fact, most existing interpretations of the soft labels are empirically and experimentally validated, while the rigorous theoretical analyses usually have strong assumptions regarding the model or data distribution.
\citet{towards} supposed that the instance could be decomposed into multiple independent features and had a linear relationship with the sample, and then prove the effectiveness of the soft labels.
\citet{linear} explored the mechanism of distillation where the teacher model and the student model are linear.
\citet{selfkdtheory} solved the objective functional problem of self-distillation with the Green's function, which assumes that the network can reach the optimal position.
\citet{statistical} and \citet{rethinking} regarded the generated soft labels as the posterior probability and assumed the existence of the Bayes probability.
Nonetheless, most of the existing theories only pertain to soft labels that are close to the ground-truth labels and lack an explicit description of the threshold condition.

This paper starts from studying biased soft labels.
To characterize the effectiveness of the soft labels, we define the top-$k$ set of the soft label, so that the problem transfer from the continuous label space to a discrete space.
Furthermore, two criteria, unreliability degree and ambiguity degree, are proposed to measure the effectiveness.
Based on the two criteria, we give the threshold conditions that guarantee the \textit{classifier-consistency} and \textit{ERM learnability} of the soft-label learners.
The results is applied to three weakly-supervised frameworks.

In section \ref{method}, our theory on the soft labels will be introduced.
To validate correctness of the theory, a heuristic approach is designed to generate such soft labels.
In section \ref{WSL}, the theory is applied to three weakly-supervised frameworks, parital label learning, learning with additive noise and learning with incomplete data.
Based on an ideal accuracy function, we further provide a coarse analysis of incomplete supervision. 
In section \ref{Experiments}, experiments on benchmark imgae datasets demonstrate that learning from biased soft labels can also achieve an adequate performance and validate that our theory is reasonable.

\section{Methodology}
\label{method}
\subsection{Preliminary}
Let $\mathcal{X}$ be the instance space and $\mathcal{Y}=\{1,2,\dots,c\}$ be the label space with $c$ classes.
The instance variable is denoted by $\boldsymbol{x} \in \mathcal{X}$ and the true label is denoted by $y \in \mathcal{Y}$.
To distinguish with the ground-truth $y$, let $\mathcal{P}$ denote the space of soft labels and $\boldsymbol{d} \in \mathcal{P}$ denote the soft label, which satisfies $\boldsymbol{d}^{i} \in [0,1]$ and $\sum_{i} \boldsymbol{d}^{i}=1$.
Here, $\boldsymbol{d}^{i}$ represents the value of $i$-th label.
$\mathcal{P}$ could be induced by prior knowledge or a teacher model $f$.
We denote the soft label of instance $\boldsymbol{x}$ by $\boldsymbol{d}_{\boldsymbol{x}}$.
The hypothesis space is denoted by $\mathcal{H}$ and each $h \in \mathcal{H}$ is a classifier (the student model) that learns from $\mathcal{X} \times \mathcal{P}$.
Note that the soft label has the same formalization as the probability distribution and, in fact, our results can also adapt to unnormalized logits.

\subsection{Theoretical Analysis}
\label{theory}
Rather than relying on criteria such as accuracy to measure the effectiveness of soft labels, reasonable indicators are necessary.
Intuitively, we refer to the rank of the true label as the measure.
First of all, we define $\Omega_{k}(\boldsymbol{d})=\{i \in \mathcal{Y} \mid i \text{ ranks top-}k \text{ in }\boldsymbol{d}\}$ as the set of top $k$ labels in the soft label $\boldsymbol{d}$.
Here, $k$ is a constant ranging in $\{1,2,\dots,c\}$.
When $k=1$, $\Omega_{k}(\boldsymbol{d})$ has only one element, i.e., the prediction.
The effectiveness of the soft labels can be reflected by whether the ground-truth label is in the $\Omega_{k}(\boldsymbol{d})$.
We define the unreliability degree $\Delta$ as,
\begin{equation} 
   \label{delta}
   \Delta=\operatorname{Pr}_{(\boldsymbol{x}, y) \sim \mathcal{X} \times \mathcal{Y}}(y \notin \Omega_{k}(\boldsymbol{d}_{\boldsymbol{x}})).
\end{equation}
Without Bayes probability, we transform the metric from the continuous label space to the discrete set of relationships, which is easy to calculate in practice.

In addition, it is not enough to measure the soft labels by merely unreliability degree.
For example, for images whose ground-truth label is $1$, if the probability that label $1$ appears in $\Omega_{k}(\boldsymbol{d})$ is inferior to label $2$, then the student model could be unable to distinguish label $2$.
Therefore, we introduce the ambiguity degree \cite{ambiguity}
\begin{equation} 
   \label{gamma}
   \gamma=\sup_{(\boldsymbol{x}, y, \boldsymbol{d}) \sim \mathcal{X} \times \mathcal{Y} \times \mathcal{P}
   \atop
   i \in \mathcal{Y}, i \neq y}
    \operatorname{Pr}(i \in \Omega_{k}(\boldsymbol{d}_{\boldsymbol{x}})).
\end{equation}
Ambiguity degree bound the probability of \textit{co-occurrence}.
In other words, if a problem exists ambiguity degree $\gamma$, then $\operatorname{Pr}(i \in \Omega_{k}(\boldsymbol{d}_{\boldsymbol{x}}) \mid i \neq y, x, y) \leq \gamma$. 
The smaller $\Delta$ or $\gamma$ is, the more supervised information the soft labels contain. 
However, when $k$ increases, $\Delta$ will decrease and $\gamma$ is inverse, which means $k$ should be selected cautiously.

\begin{theorem}
\label{E}
If $\gamma<1-\frac{\Delta}{1-\Delta}$, which means the degree of the ground-truth label is large enough, then the optimal bayesian classifier $h^{*}$ satisfies $h^{*}=\mathop{\arg\min}\limits_{h \in \mathcal{H}} R(h)$.
\end{theorem}
The proof can be found in \ref{proofE}.
Theorem \ref{E} ensures that model $h$ learning from such soft labels is optimized towards the optimal model $h^{*}$.
This property is known as classifier-consistency \cite{rccc}, which is the statistical property of the student model over the entire data distribution $\mathcal{X} \times \mathcal{Y}$. 
However, it does not provide a guarantee for models which are trained on the practical dataset and cannot ensure the convergence of the model.

Next we will prove our main result, a sufficient condition for the ERM learnability of learning with the soft labels (student model).
Firstly, we denote some common notations of machine learning theory. 
The generalization error of $h$ is defined as
\begin{equation*}
\operatorname{Err}(h)=E_{(\boldsymbol{x}, y) \sim \mathcal{X} \times \mathcal{Y}} \mathbb{I}(h(\boldsymbol{x}) \neq y).
\end{equation*}
Correspondingly, we define the soft label based generalization error and the empirical error as
\begin{equation*}
\operatorname{Err}^{\mathcal{P}}(h) =E_{(\boldsymbol{x}, y, \boldsymbol{d}) \sim \mathcal{X} \times \mathcal{Y} \times \mathcal{P}} \mathbb{I}(h(\boldsymbol{x}) \notin \Omega_{k}(\boldsymbol{d}_{\boldsymbol{x}})),
\end{equation*}
\begin{equation*}
\operatorname{Err}_{\mathbf{z}}^{\mathcal{P}}(h) =\frac{1}{n} \sum_{i=1}^n \mathbb{I}\left(h\left(\boldsymbol{x}_i\right) \notin \Omega_{k}(\boldsymbol{d}_{\boldsymbol{x}_{i}})\right),
\end{equation*}
where $\mathbf{z}$ is a dataset of size $n$.
In the above equations, the set $\Omega_{k}(\boldsymbol{d}_{\boldsymbol{x}})$ is determined by the soft label $\boldsymbol{d}_{\boldsymbol{x}}$.
We denote $H_{\Delta}$ as the set of teacher models whose generated soft labels are with unreliability degree $\Delta$, i.e., $H_{\Delta}=\{f:\operatorname{Pr}_{(\boldsymbol{x}, y, \boldsymbol{d}) \sim \mathcal{X} \times \mathcal{Y} \times \mathcal{P}}(y \notin \Omega_{k}(\boldsymbol{d}_{\boldsymbol{x}})) = \Delta\}$.
With such soft labels, the task is to train a student model that has good generalization.
We analyze the performance of the student models that minimize the empirical
risk (ERM learners).
For the hypotheses space $\mathcal{H}$ and the empirical error $\operatorname{Err}_{\mathbf{z}}^{\mathcal{P}}(h)$, the ERM learner $\mathcal{A}(\mathbf{z})$ returns the minimum empirical error on dataset $\mathbf{z}$.
\begin{equation*}
\mathcal{A}(\mathbf{z})=\mathop{\arg\min}\limits_{h \in \mathcal{H}} \operatorname{Err}_{\mathbf{z}}^{\mathcal{P}}(h).
\end{equation*}

Based on the unreliability degree $\Delta$ in (\ref{delta}) and the ambiguity degree $\gamma$ in (\ref{gamma}), we provide a sufficient condition that learning with the soft labels is ERM learnable.
The principal result is as follows.

\begin{theorem}
   \label{main}
   (Main theory) Suppose unreliability degree $\Delta$ and ambiguity degree $\gamma$, $0<\Delta, \gamma<1$ and $\Delta + \gamma<1$.
   Let $\theta=\log\frac{2(1-\Delta)}{1-\Delta+\gamma}$ and suppose the Natarajan dimension of the hypothesis space $\mathcal{H}$ is $d_\mathcal{H}$. Define
\begin{equation*}
\begin{aligned}
      n_{0}(\mathcal{H},\varepsilon,\delta)&=\frac{2}{\frac{\theta\varepsilon}{2} +\log\frac{1}{2-2\Delta}}(d_{\mathcal{H}}(\log(2d_{\mathcal{H}}) +log\frac{1}{\frac{\theta\varepsilon}{2}+\log\frac{1}{2-2\Delta}}+2\log L)+\log\frac{1}{\delta}+1).
\end{aligned}
\end{equation*}
   Then when $n>n_{0}$, the ERM learner $\mathcal{A}(\mathbf{z})$ satisfies $\operatorname{Err}(\mathcal{A}(\mathbf{z}))<\varepsilon$ with probability $1-\delta$.
\end{theorem}

We follow the methodology of proving the ERM learnability of partial label learning \cite{learnability} and the overall proof is in the appendix \ref{proofmain}. 
We define $H_{\varepsilon}$ as the set of hypotheses with error at least $\varepsilon$, i.e, $H_{\varepsilon}=\{h \in \mathcal{H}:Err(h) \geq \varepsilon\}$.
Our target is to bound the $H_{\varepsilon}$, which ensures the generalization of the learner $h$.
Since the entire soft label space is inaccessible, $H_{\varepsilon}$ is evaluated by the mediator $R_{n, \varepsilon}$ as follows:
\begin{equation*}
   R_{n, \varepsilon}=\left\{\mathbf{z} \in(\mathcal{X} \times \mathcal{Y} \times \mathcal{P})^{n}: \exists h \in H_{\varepsilon}, Err_{\mathbf{z}}^{\mathcal{P}}(h)=0\right\}.
\end{equation*}
Then, our goal is to prove that $\operatorname{Pr}(R_{n, \varepsilon}\mid f \in H_{\Delta}) \leq \delta$. 
In other words, the student model $h$, which has been trained with the generated soft labels of teacher $f$, has the generalization bound $\delta$.
Essentially, it should be clarified that the label set $\Omega_{k}(\boldsymbol{d}_{\boldsymbol{x}})$ for instance $\boldsymbol{x}$ in our proof is either induced by the teacher model or artificial generation, so the true label may not be included in $\Omega_{k}(\boldsymbol{d}_{\boldsymbol{x}})$, which is different from \citet{learnability}.

ERM learner here can be seen as selecting one confident label from the top-$k$ label set induced by the teacher model.
Different from directly optimizing the discrete loss function $Err_{\mathbf{z}}^{\mathcal{P}}(h)$, many surrogate loss functions have been proposed, of which, re-weighting is the commonest strategy.
In practice, the student model may not strictly be the ERM learner, but the essence behind it is to find the ground-truth label from the top-$k$ label set. 
The theorem can be adjusted according to the practical scenario.

Since the teacher model $f$ is intractable and the soft label space $\mathcal{P}$ is unknown, it is very difficult to directly calculate the conditional probability $\operatorname{Pr}(R_{n, \varepsilon}\mid f \in H_{\Delta})$. 
We bound it by introducing a testing set $\mathbf{z}^{\prime}$. 
The overall proof can be divided into two parts. 
Lemma \ref{le1} is used in many learnability proofs.

\begin{lemma}
   \label{le1}
   For a testing set $\mathbf{z}^{\prime} \in (\mathcal{X} \times \mathcal{Y} \times \mathcal{P})^{n}$, we can define the set $S_{n, \varepsilon}$ as
   \begin{equation*}
    \begin{aligned}
      S_{n, \varepsilon}=\{& \left(\mathbf{z}, \mathbf{z}^{\prime}\right) \in(\mathcal{X} \times \mathcal{Y} \times \mathcal{P})^{2 n}: 
      \exists h \in H_{\varepsilon}, \operatorname{Err}_{\mathbf{z}}^{\mathcal{P}}(h)=0, \operatorname{Err}_{\mathbf{z}^{\prime}}^{\mathcal{P}}(h) \geq \frac{\varepsilon}{2}\}.
    \end{aligned}
   \end{equation*}
   Then $\operatorname{Pr}((\mathbf{z},\mathbf{z}^{\prime}) \in S_{n, \varepsilon}\mid f \in H_{\Delta}) \geq \frac{1}{2}\operatorname{Pr}(\mathbf{z} \in R_{n, \varepsilon}\mid f \in H_{\Delta})$ for $n > \frac{8 \log{2}}{\varepsilon}$.
\end{lemma}


Detailed proof of Lemma \ref{le1} can be found in appendix \ref{proofle1}.
By lemma \ref{le1}, the estimation on $R_{n, \varepsilon}$ can be transformed into the estimation on $S_{n, \varepsilon}$. 
It seems more complicated but we can swap training/testing instance pairs, which is a classic method in the proof of learnability, to refine the data distribution on $\mathcal{X} \times \mathcal{Y} \times \mathcal{P}$ into each single instance. 

\begin{lemma}
   \label{le2}
   On the same condition of theorem \ref{main}.
   If the hypothesis space $\mathcal{H}$ has Natarajan dimension $d_{\mathcal{H}}$, $\gamma < 1$ and $\Delta < 1$, then 
   \begin{equation*}
      \operatorname{Pr}\left(S_{n, \varepsilon} \mid f \in H_{\Delta}\right) \leq(2 n)^{d_{\mathcal{H}}} L^{2 d_{\mathcal{H}}} \exp \left(-\frac{n \theta \varepsilon}{2}\right).
   \end{equation*}

\end{lemma}

The detailed proof of Lemma \ref{le2} can be found in appendix \ref{proofle2}.
Here, let me briefly elucidate the idea of the proof.
The fundamental technique is how to deal with the $\operatorname{Err}_{\mathbf{z}}^{\mathcal{P}}(h)$ and  $\operatorname{Err}_{\mathbf{z}^{\prime}}^{\mathcal{P}}(h)$ in $S_{n, \varepsilon}$.
Initially, we introduce the \textit{swap} $\sigma$, that swapping the instance pair of the training set $\mathbf{z}$ and testing set $\mathbf{z}^{\prime}$.
A swap $\sigma(z, z') = (z^{\sigma},z'^{\sigma})$ means exchanging some instances between the training set $z$ and testing set $z'$ while keeping size $n$ unchanged.
There are $2^{n}$ different swaps in total and we define $G$ as the set of all swaps.

Subsequently, to further refine $S_{n, \varepsilon}$, we define $S_{n, \varepsilon}^{h}$ for a certain classifier $h$ 
\begin{displaymath}
   S_{n, \varepsilon}^{h} = \{ (z,z'):\operatorname{Err}_{\mathbf{z}}^{\mathcal{P}}(h)=0, \operatorname{Err}_{\mathbf{z'}}^{\mathcal{P}}(h)\geq\frac{\varepsilon}{2} \}.
\end{displaymath}
As a result, we can bound $S_{n, \varepsilon}$ with $S_{n, \varepsilon}^{h}$ as
\begin{displaymath}
   \begin{aligned}
   & \sum_{\sigma\in G} \operatorname{Pr}(\sigma(z, z') \in S_{n, \varepsilon} \mid  x, y, x', y', f \in H_\Delta) \\
   \leq & \sum_{h \in H \mid  (x, x')} \sum_{\sigma\in G} \operatorname{Pr}(\sigma(z, z') \in S_{n, \varepsilon}^{h} \mid  x, y, x', y', f \in H_\Delta).
   \end{aligned}
\end{displaymath}
Ultimately, we separate the pair of swapped instances into three classes, both incorrectly, one incorrectly, and both correctly.
According to the numbers of each class, the condition probability $\operatorname{Pr}\left( \sigma(z, z') \in S_{n, \varepsilon}^{h} \mid  x, y, x', y', f \in H_\Delta \right)$ can be calculated precisely and the upper bound of $\sum_{\sigma\in G} \operatorname{Pr}(\sigma(z, z') \in S_{n, \varepsilon} \mid  x, y, x', y', f \in H_\Delta)$ can be estimated.



In this subsection, we establish two essential properties of the soft labels. 
Classifier-consistency guarantees the effectiveness of the soft labels in a macroscopic perspective, and ERM learnability provides a microcosmic generalization bound for the student model $h$ and the sample complexity for the realizable cases.
Furthermore, the corresponding threshold conditions are presented to ensure the student model can learn from the soft labels.
The theory is applicable to both small-bias soft labels and large-bias soft labels.
Our findings are illustrative for the comprehension and development of the soft label based algorithms.

\subsection{Customized soft labels}
\label{CSL}
In order to validate the rationality of our theory, we hope to generate an effective teacher model satisfying conditions in Theorem \ref{main} but with low accuracy. 
So we design a heuristic loss function and have some hyperparameters that qualitatively control the proposed indicators, unreliability degree and ambiguity degree.
The intuition behind the customized soft labels is to keep the ground-truth label in the top-$k$ label set but not the top.

Firstly, the teacher model will punish those correctly predicted instances as
\begin{equation}
   \mathcal{L}_\text{pun}(\boldsymbol{x}, y)=- \mathbb{I}(\mathop{\mathrm{argmax}}_{j \in \mathcal{Y}}(\boldsymbol{d}_{j})=y)\ell(f(\boldsymbol{x}), y),
\end{equation}
where $\mathbb{I}(\cdot)$ is the indicator function and $\ell(\cdot, \cdot)$ is the cross entropy loss function. 
But in practice, the value of the ground-truth label decreases significantly, resulting in large $\Delta$.
So the true label $y$ is compensated when $y \notin \Omega_{k}(\boldsymbol{d}_{\boldsymbol{x}})$:
\begin{equation}
   \mathcal{L}_\text{comp}(\boldsymbol{x}, y)= \mathbb{I}(y \notin \Omega_{k}(\boldsymbol{d}_{\boldsymbol{x}}))\ell(f(\boldsymbol{x}), y).
\end{equation}
The compensation term is designed to improve the top-$k$ accuracy of the teacher model, which keeps the statistical effectiveness of the generated soft labels.
In practice, however, we discovered there was a strong correlation among the top-$k$ labels, which leaded to the confusion between the true label and similar labels, i.e., large $\gamma$.
To decrease this correlation, we propose an effective method to make the labels in $\Omega_{k}(\boldsymbol{d}_{\boldsymbol{x}})$ as independent as possible.
Except for the true label, we randomly select $k-1$ labels. 
Then the selected $k-1$ labels are employed as the learning objectives:
\begin{equation}
   \label{srnd}
   \mathcal{L}_\text{rnd}(\boldsymbol{x}, y)=\ell(f(\boldsymbol{x}), s_\text{rnd}).
\end{equation}
where $s_\text{rnd}$ is the set of the $k-1$ random labels excluding $y$. 
Consequently, the total objective of the teacher model is as follows:
\begin{equation}
   \begin{aligned}
      \label{final}
   \mathcal{L}(\boldsymbol{x}, y)&=\mathcal{L}_\text{ce}(\boldsymbol{x}, y)+\alpha_{1}\mathcal{L}_\text{pun}(\boldsymbol{x}, y) +\alpha_{2}\mathcal{L}_\text{comp}(\boldsymbol{x}, y) +\alpha_{3}\mathcal{L}_\text{rnd}(\boldsymbol{x}, y)
   \end{aligned}
\end{equation}
where $\mathcal{L}_\text{ce}(\boldsymbol{x}, y)$ is the vanilla cross-entropy loss between the output and the ground-truth label, and $\alpha_{1}, \alpha_{2}$, $\alpha_{3}$ are the tradeoff parameters.

\section{Adaptations to weakly-supervised learning}
\label{WSL}
Soft labels are widely used in weakly-supervised learning (WSL).
The labels in WSL could be incomplete, inexact, inaccurate \cite{wsl}, because accurately labeled data is often expensive and difficult to obtain.
Due to the lack of the supervisory information, the soft labels could be large-bias but the model can still learn from them.
In this section, our theory is adapted to three classic weakly-supervised frameworks and can elucidate the feasibility behind them.
These findings reflect that the theory is promising and extensible.


\subsection{Partial label learning} In partial label learning (PLL), the ground-truth label of each instance $y$ is replaced by the candicate label set $s$ \cite{Progressivepll, rcr, idpll}.
The corresponding candicate label space is denoted by $\mathcal{S}$.
Traditional PLL assumes the ground-truth label must be in the candicate label set but recently, \citet{upll} considers that ground-truth label could be not in the candicate label set, which is named as
unreliable PLL (UPLL).

In UPLL, there are two basic concepts, partial rate $\eta$ and unreliable rate $\mu$.
Partial rate $\eta$ is the ratio of incorrect labels in the candicate label set to total labels.
A lower partial rate usually indicates a better performance of the model.
Unreliable rate $\mu$ is the probability of ground-truth label $y$ not in the candidate label set, which can be formally stated as
\begin{equation*}
   \mu=\operatorname{Pr}_{(\boldsymbol{x}, y, s) \sim \mathcal{X} \times \mathcal{Y} \times \mathcal{S}}(y \notin s).
\end{equation*}

The discrete candicate label set $s$ can be transformed into the soft label by 
\begin{equation*}
   d_i = \left\{
      \begin{aligned}
      \frac{1}{|s|} \quad i \in s\\
      0 \quad i \notin s\\
      \end{aligned}
      \right.,
\end{equation*}
where $|s|$ is the cardinality of set $s$.
So PLL also can be viewed as learning from soft labels.
Then we have the following corollary.

\begin{corollary}
   For UPLL with partial rate $\eta$ and unreliable rate $\mu$, 
   we have $\Delta=\mu$ and $\gamma=\eta$.
   With the same conditions in Theorem \ref{main}, UPLL is ERM learnable and the sample complexity remains unchanged.
   
\end{corollary}

Most algorithms for PLL and UPLL re-weight the loss by the outputs of the model.
It is instructive to understand the PLL from the perspective of soft labels. 

\subsection{Learning with additive noise} 
Additive noise mechanism \cite{an, an2} is an important methodology for differential privacy.
Specifically, Laplace noise or Gaussian noise is added to data for protecting privacy.
The privacy budget can be controlled by adjusting the scale of the noise.
After normalization, the noisy labels are also soft labels in nature.
In fact, given the probability density function of noise, we can calculate the corresponding unreliability degree and ambiguity degree in order to measure the effectiveness of the noisy labels.
Based on the noisy labels, the task is to train a utility model with strong privacy guarantees.
Our theory can guarantee the utility of such soft labels.

To depict labels with additive noise, we refer to \textit{order statistic} \cite{order}.
Order statistic analyze the $i$th-smallest value of random samples from a continuous distribution.
We denote the \textit{order distribution} $Order(d, n, i)$ as the $i$th-smallest value of $n$ samples from distribution $d$.
The software Mathematics \cite{mathematics} provide an efficient API for estimating the order distribution.

\begin{corollary}

   
   Let $d$ denote the noise distribution (e.g. Laplace noise and Gaussian noise).
   With the $k$ in Eq.\ref{delta} and the total classes $c$, for $k \leq c-1$, we can compute the $\Delta$ and $\gamma$ as
   \begin{equation*} 
      \Delta=\operatorname{Pr}_{x \sim Order(d, c-1, n-k+1)
      \atop
      y \sim d
      }(1+y>x),
   \end{equation*}

   \begin{equation*} 
      \gamma=\frac{\Delta+k-1}{c-1}.
   \end{equation*}
   With the same conditions in Theorem \ref{main}, learning with additive noise is ERM learnable and the sample complexity remains unchanged.
\end{corollary}

As the scale of the nosie increases, $\Delta$ and $\gamma$ will increase, i.e., the effectiveness of the soft labels will decrease.
This result agrees with the practical situation.

\subsection{Learning with incomplete data} 
\label{4.3}
Many real-world applications lack sufficient labeled data due to cost, but have a great deal of unlabeled data \cite{unlabel1, unlabel2, unlabel3}.
Incomplete supervision is proposed to reduce the amount of time and resources needed to train a deep learning model.
A common approach is to use the predictive model to label the unlabeled data and then learn with all data iteratively.
This progress can be viewed as a variant of self-distillation that the model teaches itself.
The soft labels of the unlabeled data are thought to contain much supervisory information and play a crucial role in this progression.

Here, disregarding the model architecture, data distribution and optimization, we propose a coarse analysis on learning with incomplete data in the light of soft labels and our prior theory.
Suppose there are $N$ labeled data and $M$ unlabeled data sampled from $\mathcal{X} \times \mathcal{Y}$.
The predictive model $h$ is an ERM learner on both labeled data and unlabeled data.
The label of unlabeled data will be updated iteratively.

\begin{assumption}
   For $N$ labeled data and $M$ unlabeled data whose soft labels have unreliability degree $\Delta$ and ambiguity degree $\gamma$, the ERM learner $h$ has a deterministic accuracy funtion $\rho(\Delta, \gamma)$, the probability that $h$ predict correctly.
   The model architecture, data distribution and optimization are implicitly included in $\rho(\Delta, \gamma)$.
\end{assumption}

\begin{assumption}
   Since that the smaller $\Delta$ or $\gamma$ is, the more supervised information is in the soft labels, we assume $\rho(\Delta, \gamma)$ decreases with $\Delta$ and $\gamma$.
\end{assumption}

We suppose that the noise in the soft labels is uniformly distributed, implying that the incorrect labels share the equal probability.
More intricate condition for the noise distribution can be formulated.
For instance, there is a upper bound of $\frac{p(i|\boldsymbol{x})}{p(j|\boldsymbol{x})}, i,j \in \mathcal{Y}, i,j \neq y, i \neq j$.
Simplified assumptions can also reflect this process.

Based on the ideal $\rho(\Delta, \gamma)$, we delineate the progressive performance of $h$ as
\begin{equation}
   \label{iterative}
   \rho_{t+1} \geq \rho(1-\rho_{t}, \frac{c-k-\rho_{t}}{c-1}).
\end{equation}
where $\rho_{t}$ is the accuracy of $h$ at epoch $t$.
In practice, as learner $h$ learns from labeled data and unlabeled data, the performance of $h$ will improve and the soft labels of the unlabeled data will be more effective.
Consequently, $h$ and the soft labels may achieve a dynamic equilibrium.
Specifically, if $\rho(\Delta,\gamma)$ is $k_{L}$-\textit{Lipschitz} continuous ($k_{L}<1-\frac{1}{c}$), the accuracy of $h$ will reach a moderate level and the final accuracy $\rho_{\text{final}}$ can be calculated by the fixed point equation.

\begin{theorem}
   \label{rho}
   Based on the ideal accuracy function $\rho(\Delta, \gamma)$, with a moderate initial state $\Delta_0, \gamma_0$ satisfying Theorem \ref{main}, if final accuracy of $\rho_{\text{final}}$ exists, it can be calculated by the following fixed point equation:
   \begin{equation}
      \label{fix}
      x = \rho(1-x, \frac{c-k-x}{c-1}).
   \end{equation}
   where $k$ accords with the top-$k$ set in $\Delta$ and $\gamma$, $c$ is the number of class labels.
   If $\rho(\Delta,\gamma)$ is $k_{L}$-\textit{Lipschitz} continuous ($k_{L}<1-\frac{1}{c}$), then $\rho_{\text{final}}$ exists and is unique.
\end{theorem}



The proof is detailed in \ref{proofrho}.
In fact, the deterministic $\rho(\Delta, \gamma)$ is unattainable due to the indeterminacy of the optimization and the potential uncertainty in the data distribution.
An intuitive extension is to assume $\rho(\Delta, \gamma)$ is a probability distribution related to the training specifics, which can be further investigated.
Theorem \ref{rho} is coarse yet in agreement with the general intuition.
The model $h$ improves as the soft labels envolve and finally reach the bottleneck restricted by the model, data and optimization.

In this section, we demonstrate the potential benefits of our theory in several classic weakly-supervised frameworks.
There remain numerous domains associated with soft labels.
It is essential to possess an appropriate theory to analyze the soft labels for the corresponding algorithms.
Our theory can be instrumental for comprehending and constructing the soft label based algorithms.

\section{Experiments}
\label{Experiments}

\subsection{Experiment Setup}
We consider two benchmark imgae datasets CIFAR-10 and CIFAR-100 \cite{CIFAR} and generate the soft labels with different hyperparameters of the teacher model.
The student model is trained with the modified labels and aims to distinguish the true label. 
The accuracy of the student model is employed to measure the effectiveness of the soft labels.
Datasets are divided into training, validation, testing set in the ratio of 4:1:1.
For the fairness of the experiments, all student models are WideResNet28$\times$2 architecture \cite{wrn} on each dataset.
In all experiments, we use mini-batch SGD \cite{sgd} with a batch size of 128 and a momentum of 0.9.
Each model is trained with maximum epochs $T=200$ and employs early stopping strategy with patience 20.
Finally, we report final performance using the test accuracy corresponding to the best accuracy on validation set.
For the following experiments, we employ a basic re-weighting strategy to train the student model.
Specifically, in each epoch, we train the student model and update the soft labels with the softmax outputs of it.

\begin{figure*}[t]
   \centering
   \subfigure[]{
   \begin{minipage}[t]{0.5\linewidth}
   \centering
   \hspace{-0.0cm}\includegraphics[width=2.95in]{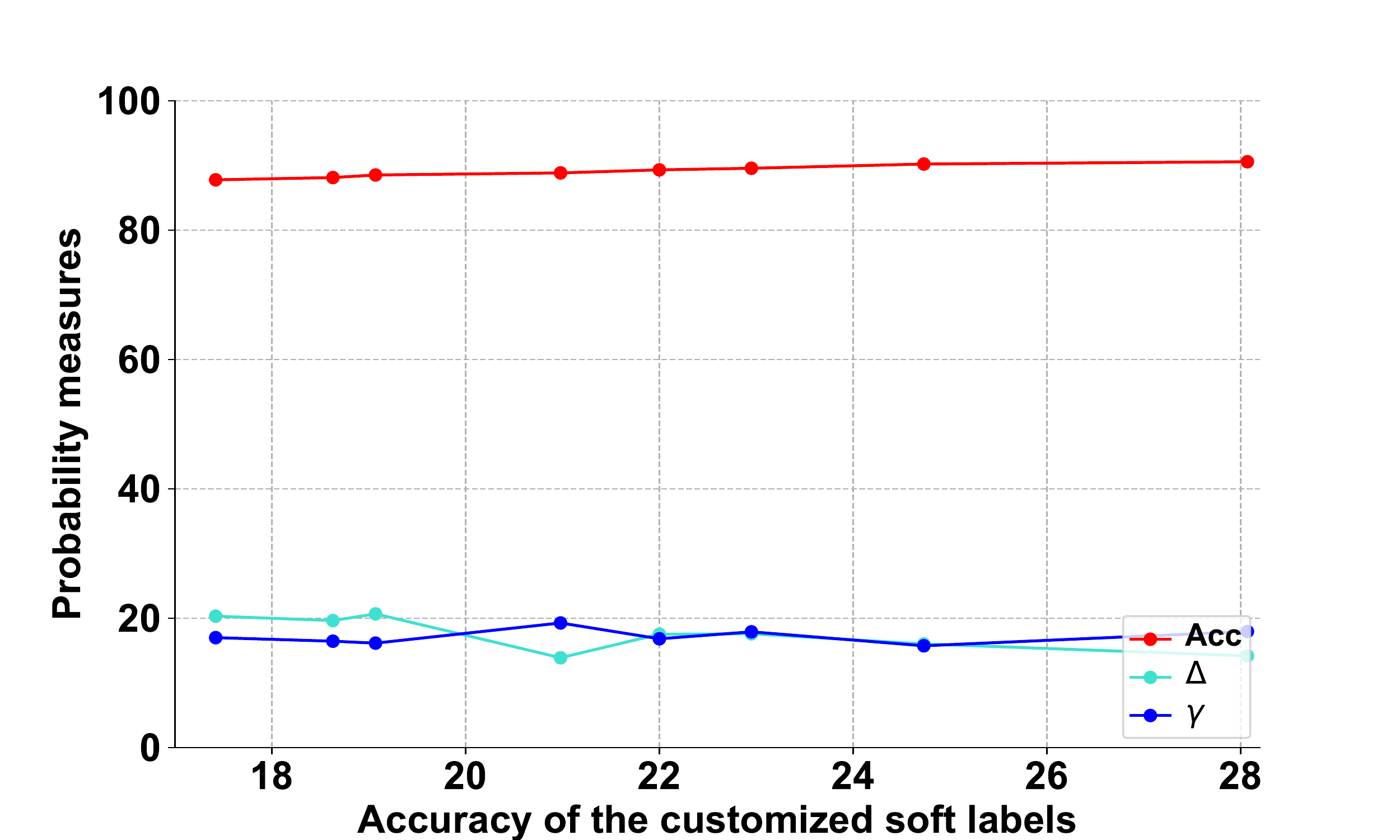}
   \end{minipage}%
   }%
   \subfigure[]{
   \begin{minipage}[t]{0.5\linewidth}
   \centering
   \includegraphics[width=2.95in]{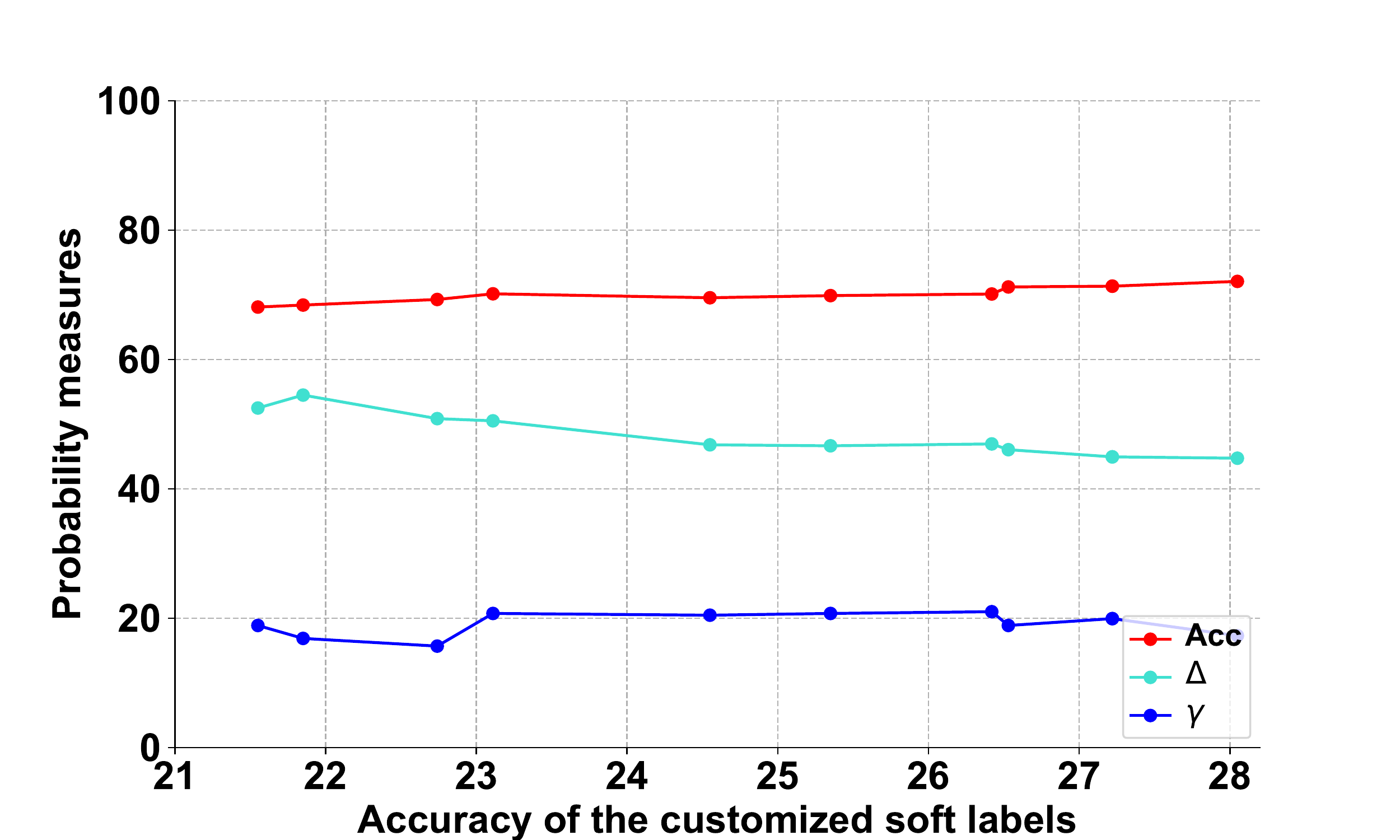}
   \end{minipage}%
   }%
   \centering
   \caption{Indicators of the customized soft labels on CIFAR-10 (a) and CIFAR-100 (b).}
   \label{pu}
\end{figure*}

\subsection{Effectiveness of the proposed indicators}
We propose two indicators in section \ref{method}, unreliability degree $\Delta$ and ambiguity degree $\gamma$, to measure the effectiveness of the soft labels.
Furthermore, although the soft labels are large-bias (i.e., with low accuracy), the indicators still work and Theorem \ref{main} provides a guarantee for the learners (students).
We illustrate this phenomenon with the customized soft labels introduced in \ref{CSL} as Figure \ref{pu}.

There are three probability measures, Acc($\uparrow$), $\Delta$($\downarrow$) and $\gamma$($\downarrow$) to quantify the effectiveness of the soft labels.
Acc refers to the accuracy of the student model that learns from the soft labels.
The Acc can be considered as \textit{the ground-truth effectiveness of the soft labels}.
On the hand, $\Delta$ and $\gamma$ are the direct measures without training students.
Note that training with the ground-truth labels can achieve 95.29\% on CIFAR-10 and 78.13\% on CIFAR-100. 
We can find many amazing results in Figure \ref{pu}:
\begin{itemize}
   \item The customized soft labels are with accuracy less than 30\%, which means they are quite different from the ground-truth labels, but the students achieve an adequate performance.
   \item As the accuracy of the customized soft labels increases, $\Delta$ and $\gamma$ change slightly on CIFAR-10, which is consistent with the Acc. While for CIFAR-100, $\Delta$ decreases and Acc increases.
   \item For the more complicated CIFAR-100, the soft labels generated by the teacher model could be more unreliable. $\Delta$ is relatively high and Acc is relatively low.
\end{itemize}

The hyperparameters in Eq.(\ref{final}) are pivotal in regulating the indicators of the soft labels, which is elucidated in the appendix \ref{hyper}.
In addition, we show the overall distribution of the customized soft labels in \ref{validate_2}.
The experimental results demonstrate the effectiveness of the proposed indicators and that \textit{learning from large-bias soft labels can also yield an adequate performance}.
This further corroborates the veracity of our theory.


\begin{figure*}[t]
   \centering
   \subfigure[]{
   \begin{minipage}[t]{0.5\linewidth}
   \centering
   \hspace{-0.0cm}\includegraphics[width=2.90in]{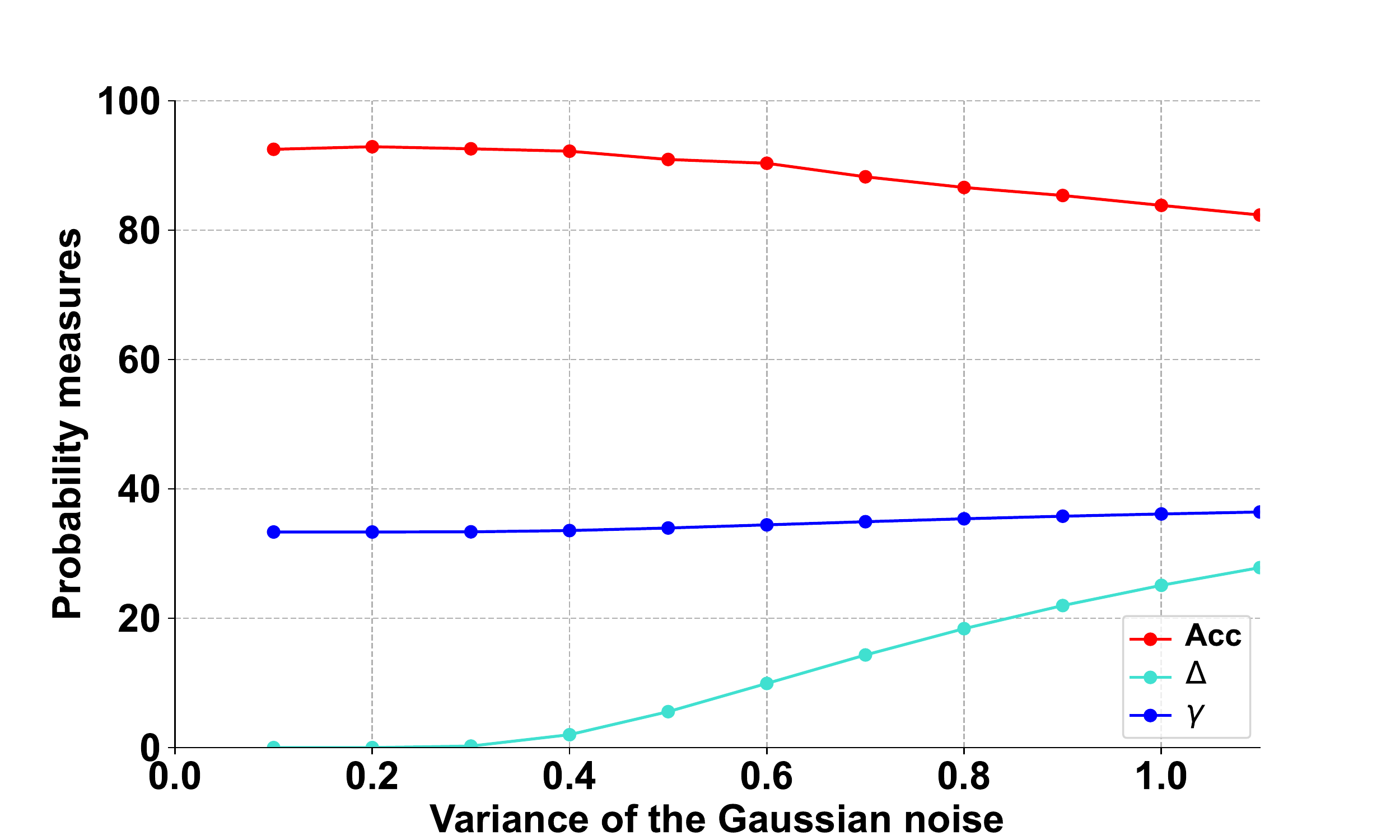}
   \end{minipage}%
   }%
   \subfigure[]{
   \begin{minipage}[t]{0.5\linewidth}
   \centering
   \includegraphics[width=2.90in]{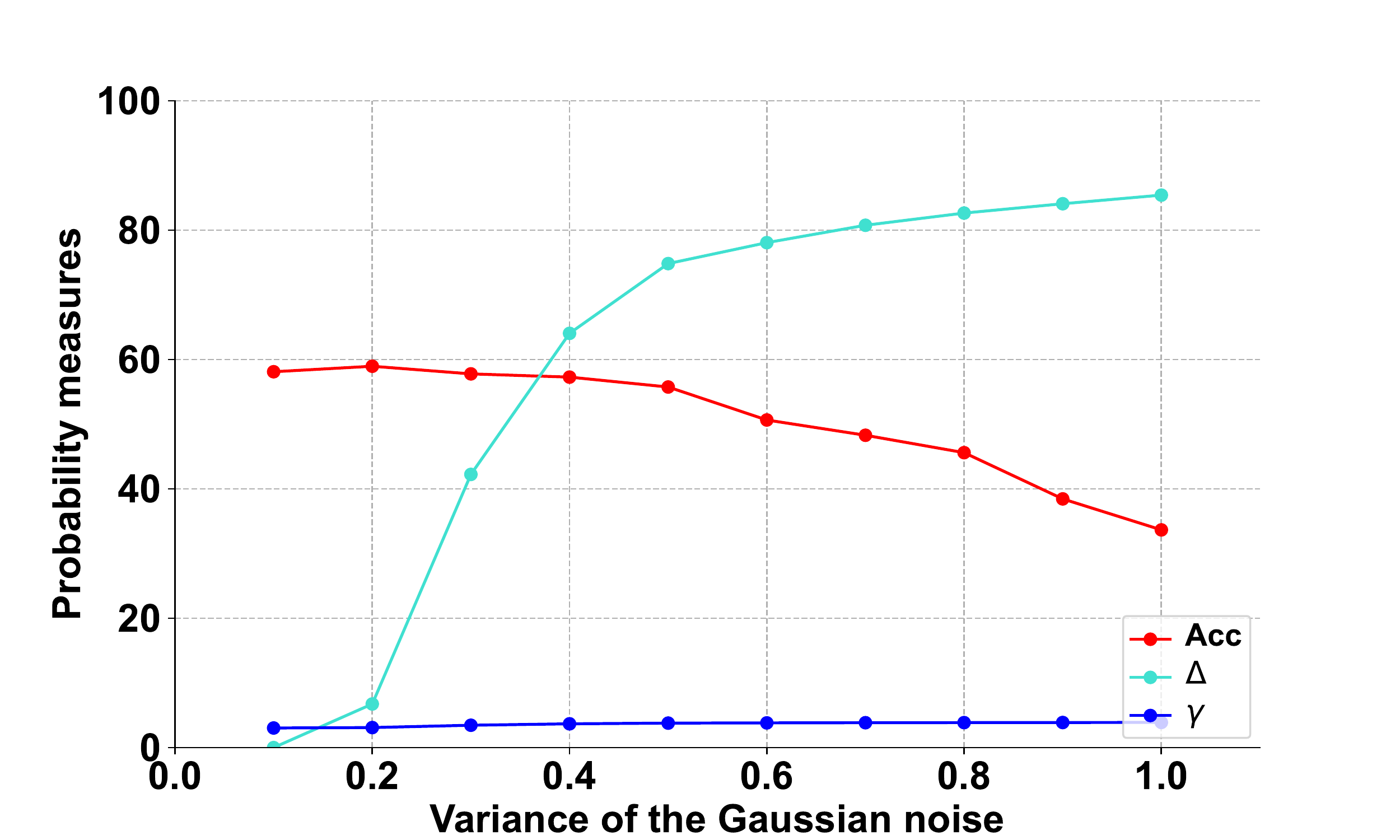}
   \end{minipage}%
   }%
   \centering
   \caption{Indicators of labels with Gaussian noise on CIFAR-10 (a) and CIFAR-100 (b).}
   \label{noise}
\end{figure*}

\subsection{Soft labels for weakly-supervised learning}
The aforesaid weakly-supervised learning paradigms have engendered a lot of specialized algorithms. 
The intention of this paper is not to devise more efficacious algorithms, but to evaluate the effectiveness of the soft labels in the weakly-supervised domain.
Constrained by space, we present the results of PLL and learning with additive noise in Table \ref{PLL} (shown in appendix) and Figure \ref{noise}, while the results of learning with incomplete data are illustrated in the appendix \ref{unlabeled}.
More details regarding to the experiments of weakly-supervised learning can be found in \ref{dwsl}.
From Figure \ref{noise} and Table \ref{PLL}, we can observe that the performance of the students decrease when $\Delta$ or $\gamma$ increase.
All the results are consistent with our theory and our theory provides a guarantee for these paradigms.


\section{Conclusion}
In this work, we focus on the effectiveness of the biased soft labels and discover that even large-bias soft labels can teach a good student. 
This phenomenon motivate us to rethink when the biased soft labels (teachers) are effective.
Base on the proposed indicators, we provide sufficient conditions that guarantee the classifier-consistency and ERM learnability of the soft-label learners, which can also be applied to weakly-supervised frameworks.
Finally, the experimental results validate that students can learn from the biased soft labels, which is consistent with our theory.

\bibliography{main}
\bibliographystyle{ims}

\newpage
\appendix

\section{Appendix}
\subsection{Proof of Theorem \ref{E}}
\label{proofE}
\begin{displaymath}
   \begin{aligned}
   R(h)= & E_{(\boldsymbol{x}, d) \sim \mathcal{X} \times \mathcal{P}}[\min_{i \in \Omega_{k}(\boldsymbol{d}_{\boldsymbol{x}})} \ell(h(x),i)\cdot p(i \in \Omega_{k}(\boldsymbol{d}_{\boldsymbol{x}})\mid \boldsymbol{d}_{\boldsymbol{x}})]
   \\
   = & E_{(\boldsymbol{x}, d) \sim \mathcal{X} \times \mathcal{P}}[\min_{i \in \Omega_{k}(\boldsymbol{d}_{\boldsymbol{x}})} \{ \ell(h(x),i)\cdot p(i \in \Omega_{k}(\boldsymbol{d}_{\boldsymbol{x}})\mid y \in \Omega_{k}(\boldsymbol{d}_{\boldsymbol{x}})) \cdot p(y \in \Omega_{k}(\boldsymbol{d}_{\boldsymbol{x}}) \mid \boldsymbol{d}_{\boldsymbol{x}})
   \\
   & +\ell(h(x),i)\cdot p(i \in \Omega_{k}(\boldsymbol{d}_{\boldsymbol{x}})\mid y \notin \Omega_{k}(\boldsymbol{d}_{\boldsymbol{x}})) \cdot p(y \notin \Omega_{k}(\boldsymbol{d}_{\boldsymbol{x}}) \mid \boldsymbol{d}_{\boldsymbol{x}}) \}].
\end{aligned}
\end{displaymath}
The coefficient of $\ell(h(x),y)$ is:
\begin{displaymath}
   \begin{aligned}
      \text{Coff}[\ell(h(x),y)] = & p(y \in \Omega_{k}(\boldsymbol{d}_{\boldsymbol{x}})\mid y \in \Omega_{k}(\boldsymbol{d}_{\boldsymbol{x}})) \cdot p(y \in \Omega_{k}(\boldsymbol{d}_{\boldsymbol{x}}) \mid \boldsymbol{d}_{\boldsymbol{x}})
      \\
      = & 1-\Delta.
   \end{aligned}
\end{displaymath}
For $i\neq y$, there is 
\begin{displaymath}
   \begin{aligned}
      \text{Coff}[\ell(h(x),i)] = & p(i \in \Omega_{k}(\boldsymbol{d}_{\boldsymbol{x}})\mid y \in \Omega_{k}(\boldsymbol{d}_{\boldsymbol{x}})) \cdot p(y \in \Omega_{k}(\boldsymbol{d}_{\boldsymbol{x}}) \mid \boldsymbol{d}_{\boldsymbol{x}})
      \\
      & + p(i \in \Omega_{k}(\boldsymbol{d}_{\boldsymbol{x}})\mid y \notin \Omega_{k}(\boldsymbol{d}_{\boldsymbol{x}})) \cdot p(y \notin \Omega_{k}(\boldsymbol{d}_{\boldsymbol{x}}) \mid \boldsymbol{d}_{\boldsymbol{x}})
      \\
      \leq & \gamma(1-\Delta)+\Delta.
   \end{aligned}
\end{displaymath}
Therefore, when $1-\Delta>\gamma(1-\Delta)+\Delta$, i.e., $\gamma<1-\frac{\Delta}{1-\Delta}$, we have $h^{*}=\mathop{\arg\min}\limits_{h \in \mathcal{H}} R(h)$.

\subsection{Proof of Lemma \ref{le1}}
\label{proofle1}
Lemma \ref{le1} is a trick widely used in the proof of learnability.
Consider the training set $z$, the testing set $z'$ and each of them is of size $n$.
Define $H(z)=\{h \in \mathcal{H}:Err_{z}^{\mathcal{P}}(h) = 0\}$ as the set of zero-empirical-risk hypotheses.
We can bound $\operatorname{Pr}\left( z \in R_{n, \varepsilon} \mid  f \in H_\Delta \right)$ with $\operatorname{Pr} \left( (z, z') \in S_{n, \varepsilon} \mid  f \in H_{\Delta} \right)$ as follows
\begin{displaymath}
\begin{aligned}
   & \operatorname{Pr} \left( (z, z') \in S_{n, \varepsilon} \mid  f \in H_{\Delta} \right)
      \\
      = & \operatorname{Pr} \left( (z, z') \in S_{n, \varepsilon} \mid  z \in R_{n, \varepsilon}, f \in H_{\Delta} \right)
      \\
      = & \operatorname{Pr} \left( \{\exists h \in H_\varepsilon \cap H(z), Err_{z'}(h) \geq \frac{\varepsilon}{2} \} \mid  z \in R_{n, \varepsilon}, f \in H_{\Delta} \right)
      \\
      \geq & \operatorname{Pr} \left(  h \in H_\varepsilon \cap H(z), Err_{z'}(h) \geq \frac{\varepsilon}{2} \mid  z \in R_{n, \varepsilon}, f \in H_{\Delta} \right)
      \\
      \geq & 1-\exp{(-\cdot\frac{\varepsilon n}{8})}.
\end{aligned}
\end{displaymath}
When $n > \frac{8 \log{2}}{\varepsilon}$, we have $\operatorname{Pr} \left( (z, z') \in S_{n, \varepsilon} \mid  f \in H_{\Delta} \right) \geq \frac{1}{2} \operatorname{Pr}\left( z \in R_{n, \varepsilon} \mid  f \in H_\Delta \right)$, which completes the proof.


\subsection{Proof of Lemma \ref{le2}}
\label{proofle2}
Here we need to bound $\operatorname{Pr}\left( S_{n, \varepsilon} \mid  f \in H_\Delta \right)$.
The key behind the proof is to refine $Err_{z}^{\mathcal{P}}(h)$ and $Err_{z'}^{\mathcal{P}}(h)$.
We use a classic method, i.e., swap, to refine the single instance.
A swap $\sigma(z, z') = (z^{\sigma},z'^{\sigma})$ means exchanging some instances between the training set $z$ and testing set $z'$ while keeping size $n$ unchanged.
There are $2^{n}$ different swaps in total and we define $G$ as the set of all swaps.
Firstly, we use swap to describe $\operatorname{Pr}\left( S_{n, \varepsilon} \mid  f \in H_\Delta \right)$.

\begin{displaymath}
\begin{aligned}
   2^n\operatorname{Pr}\left( S_{n, \varepsilon} \mid  f \in H_\Delta \right)
   = & \sum_{\sigma\in G}E\left[ \operatorname{Pr}((z, z') \in S_{n, \varepsilon} \mid  x, y, x', y', f \in H_\Delta) \right]
   \\
   = & \sum_{\sigma\in G}E\left[ \operatorname{Pr}(\sigma(z, z') \in S_{n, \varepsilon} \mid  x, y, x', y', f \in H_\Delta) \right]
   \\
   = & E\left[ \sum_{\sigma\in G} \operatorname{Pr}(\sigma(z, z') \in S_{n, \varepsilon} \mid  x, y, x', y', f \in H_\Delta) \right].
\end{aligned}
\end{displaymath}

To further refine $S_{n, \varepsilon}$, we define $S_{n, \varepsilon}^{h}$ for a certain classifier $h$ as
\begin{displaymath}
   S_{n, \varepsilon}^{h} = \{ (z,z'): Err_{z}^{\mathcal{P}}(h)=0, Err_{z'}(h)\geq\frac{\varepsilon}{2} \}.
\end{displaymath}

Next, we have the bound 
\begin{displaymath}
   \sum_{\sigma\in G} \operatorname{Pr}(\sigma(z, z') \in S_{n, \varepsilon} \mid  x, y, x', y', f \in H_\Delta) \leq \sum_{h \in H \mid  (x, x')} \sum_{\sigma\in G} \operatorname{Pr}(\sigma(z, z') \in S_{n, \varepsilon}^{h} \mid  x, y, x', y', f \in H_\Delta).
\end{displaymath}

By \cite{olsaf}, the hypotheses space $H\mid (x, x')$ can be bounded as
\begin{displaymath}
   \left|H\mid (x, x')\right| \leq (2n)^{d_H}L^{2d_H}.
\end{displaymath}

Then, 
\begin{displaymath}
\begin{aligned}
   & \operatorname{Pr}\left( \sigma(z, z') \in S_{n, \varepsilon}^{h} \mid  x, y, x', y', f \in H_\Delta \right)
   \\
   = & I\left(Err_{z'}\sigma(h)\geq\frac{\varepsilon}{2}\mid f\in H_\Delta\right)\cdot \operatorname{Pr}\left(h(x_{i}^{\sigma})\in S_i^{\sigma}, 1 \leq i \leq n \mid  x^{\sigma}, y^{\sigma}, f \in H_\Delta\right)
   \\
   = & I\left(Err_{z'}\sigma(h)\geq\frac{\varepsilon}{2}\mid f\in H_\Delta\right)\cdot \prod_{i=1}^{n} \operatorname{Pr}\left(h(x_{i}^{\sigma})\in S_i^{\sigma} \mid  x^{\sigma}, y^{\sigma}, f \in H_\Delta\right).
\end{aligned}
\end{displaymath}

For the pair of $(x, y, x',y')$, we consider the number of instances of all cases.
Specifically, let $u_{1}$, $u_{2}$ and $u_{3}$ represent the number of both incorrectly predicted instances, one incorrectly predicted instances and both correctly predicted instances.
Besides, we define $u_{\sigma}$ as the number of instances where $(x^{\sigma}, y^{\sigma})$ is incorrectly predicted while $(x'^{\sigma},y'^{\sigma})$ is correctly predicted.
Afterwards, the number of incorrectly predicted instances in the tesging set is $u_{1}+u_{2}-u_{\sigma}$.
\begin{displaymath}
   \begin{aligned}
      I\left(Err_{z'}\sigma(h)\geq\frac{\varepsilon}{2}\mid f\in H_\Delta\right)
      = & I(u_{1}+u_{2}-u_{\sigma} \geq \frac{\varepsilon}{2}n)
      \\
      \leq & I(u_{1}+u_{2} \geq \frac{\varepsilon}{2}n).
   \end{aligned}
\end{displaymath}

For $\operatorname{Pr}\left(h(x_{i}^{\sigma})\in S_i^{\sigma} \mid  x^{\sigma}, y^{\sigma}, f \in H_\Delta\right)$, we count instances which have been swapped.
On the one side, there are $u_{2}+u_{3}-u_{\sigma}$ instances satisfying $h(x_{i}^{\sigma})=y_{i}^{\sigma}$ where we have $\operatorname{Pr}\left(h(x_{i}^{\sigma})\in S_i^{\sigma} \mid  f \in H_\Delta\right) = 1 - \Delta$.
On the other side, there are $u_{1}+u_{\sigma}$ instances satisfying $h(x_{i}^{\sigma}) \neq y_{i}^{\sigma}$ where we have $\operatorname{Pr}\left(h(x_{i}^{\sigma})\in S_i^{\sigma} \mid  f \in H_\Delta\right) \leq \gamma$.
So, for any $i$, we have
\begin{displaymath}
   \begin{aligned}
   \operatorname{Pr}\left(h(x_{i}^{\sigma})\in S_i^{\sigma} \mid  x^{\sigma}, y^{\sigma}, f \in H_\Delta\right)\leq (1-\Delta)^{u_2+u_3-u_{\sigma}}\cdot\gamma^{u_1+u_{\sigma}}.
   \end{aligned}
\end{displaymath}

And then, the conditional probability can be bounded as follow:
\begin{displaymath}
\begin{aligned}
\operatorname{Pr}\left( \sigma(z, z') \in S_{n, \varepsilon}^{h} \mid  x, y, x', y'\right) \leq I\left(u_1+u_2 \geq \frac{\varepsilon}{2}n\right)(1-\Delta)^{u_2+u_3-u_{\sigma}}\cdot\gamma^{u_1+u_{\sigma}}.
\end{aligned}
\end{displaymath}

There are $2^{n}$ different swaps and we sum all.
\begin{displaymath}
   \begin{aligned}
   & \sum_{\sigma\in G}I\left(u_1+u_2 \geq \frac{\varepsilon}{2}n\right)(1-\Delta)^{u_2+u_3-u_{\sigma}}\cdot\gamma^{u_1+u_{\sigma}}
   \\
   \leq & 2^{u_1+u_3} \cdot I\left(u_1+u_2 \geq \frac{\varepsilon}{2}n\right)\cdot \sum_{j=0}^{u_2}\left(\begin{matrix}
      u_2 \\
      j
   \end{matrix}\right)(1-\Delta)^{u_2+u_3-j}\cdot \gamma^{u_1+j}
   \\
   = & 2^{n-u_2}\cdot(1-\Delta)^{u_2+u_3}\cdot\gamma^{u_1}\cdot I\left(u_1+u_2 \geq \frac{\varepsilon}{2}n\right)\cdot \sum_{j=0}^{u_2}\left(\begin{matrix}
      u_2 \\
      j
   \end{matrix}\right)(\frac{\gamma}{1-\Delta})^j
   \\
   = & 2^{n-u_2}\cdot(1-\Delta)^{n-u_1}\cdot\gamma^{u_1}\cdot I\left(u_1+u_2 \geq \frac{\varepsilon}{2}n\right)\cdot (1+\frac{\gamma}{1-\Delta})^{u_2}
   \\
   = & I\left(u_1+u_2 \geq \frac{\varepsilon}{2}n\right)\cdot 2^{n-u_2} \cdot (1-\Delta)^{n-u_1}\cdot \gamma^{u_1} \cdot (\frac{1-\Delta+\gamma}{1-\Delta})^{u_2}
   \\
   = & I\left(u_1+u_2 \geq \frac{\varepsilon}{2}n\right)\cdot  (2-2\Delta)^n\cdot (\frac{\gamma}{1-\Delta})^{u_1}\cdot (\frac{1-\Delta+\gamma}{2(1-\Delta)})^{u_2}.
\end{aligned}
\end{displaymath}
where $n=u_1+u_2+u_3$. The $2^{u_1+u_3}$ in the first step means ways of swapping both correct instances and both incorrect instances.
The index of $j$ equals $u_2$ different swaps of one correctly predicted instances.
According to the assumption $\Delta+\gamma \leq 1$, we have $0<\frac{\gamma}{1-\Delta} < \frac{1-\Delta+\gamma}{2(1-\Delta)}<1$.
For $u_1+u_2 \geq \frac{\varepsilon}{2}n$, when $u_{1}=0$ and $u_{2}=\frac{\varepsilon}{2}n$, the right side reaches its maximum as follows:
\begin{displaymath}
   \begin{aligned}
   \operatorname{Pr}\left(S_{n, \varepsilon}\mid f\in H_\Delta\right) \leq (2n)^{d_H}\cdot L^{2d_H}\cdot (2-2\Delta)^n\cdot(\frac{1-\Delta+\gamma}{2(1-\Delta)})^{\frac{n\varepsilon}{2}}.
\end{aligned}
\end{displaymath}
We have proved the Lemma \ref{le2}.

\subsection{Proof of Theorem \ref{main}}
\label{proofmain}
With Lemma \ref{le1} and Lemma \ref{le2}, we have
\begin{displaymath}
   \begin{aligned}
   \operatorname{Pr}\left(R_{n, \varepsilon}\mid f\in H_\Delta\right) \leq 2^{d_H+1}\cdot n^{d_H} \cdot L^{2d_H} \cdot (2-2\Delta)^n\cdot(\frac{1-\Delta+\gamma}{2(1-\Delta)})^{\frac{n\varepsilon}{2}}.
\end{aligned}
\end{displaymath}


We set $\theta$ as 
\begin{displaymath}
   \theta = \log{\frac{2(1-\Delta)}{1-\Delta+\gamma}}.
\end{displaymath}
Since $\Delta+\gamma<1$, we get $\theta>0$.
We need to bound $\operatorname{Pr}\left(R_{n, \varepsilon}\mid f\in H_\Delta\right)$ with $\delta$, which means
\begin{displaymath}
   (d_H+1)\cdot\log{2}+d_H\log{n}+2d_H\log{L}+n\log{(2-2\Delta)}-\frac{\theta\varepsilon n}{2} \leq \log{\delta}.
\end{displaymath}

Note that the function $f(x)=log\frac{1}{a}+ax-logx-1 \geq 0$. Let $a=\frac{\frac{\theta\varepsilon}{2}+log(\frac{1}{2-2\Delta})}{d_H}$ and $x=n$. It can be inferred that 
\begin{displaymath}
   log{n}\leq \frac{\frac{\theta\varepsilon}{2}+log(\frac{1}{2-2\Delta})}{d_H} n -log{\frac{\frac{\theta\varepsilon}{2}+log(\frac{1}{2-2\Delta})}{d_H}} -1.
\end{displaymath}
With the bound of $log{n}$, we get the linear inequality of $n$. Let
\begin{displaymath}
   n_{0}(\mathcal{H},\varepsilon,\delta)=\frac{2}{\frac{\theta\varepsilon}{2}+\log\frac{1}{2-2\Delta}}(d_{\mathcal{H}}(\log(2d_{\mathcal{H}})+\log\frac{1}{\frac{\theta\varepsilon}{2}+\log\frac{1}{2-2\Delta}}+2\log L)+\log\frac{1}{\delta}+1).
\end{displaymath}
When $n>n_{0}$, we get $\operatorname{Pr}\left(R_{n, \varepsilon}\mid f\in H_\Delta\right)<\delta$ and the proof is finished.

\begin{figure}
   \centering
   \subfigure[]{
   \begin{minipage}[t]{0.5\linewidth}
   \centering
   \hspace{-0.2cm}\includegraphics[width=2.5in]{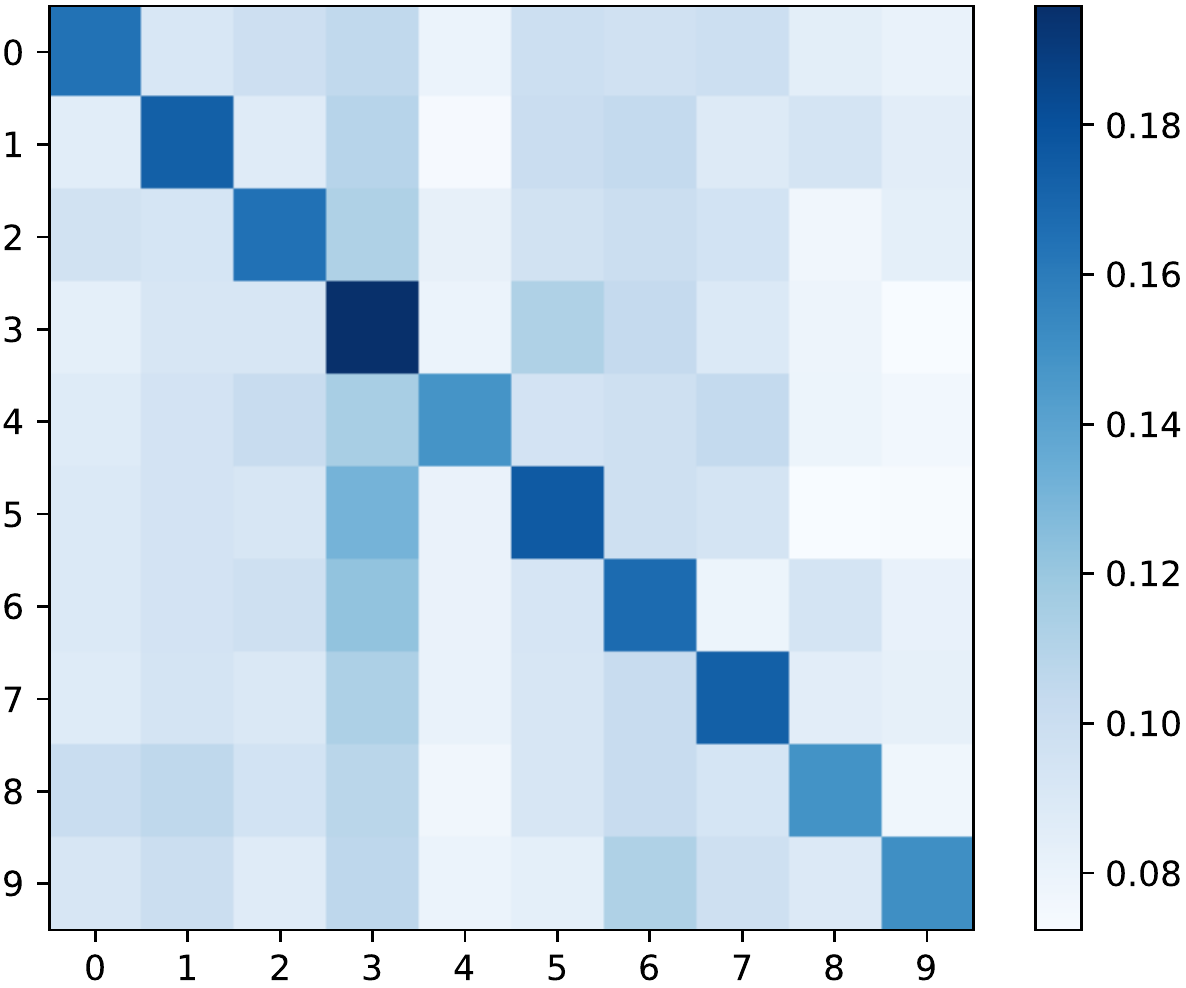}
   \end{minipage}%
   }%
   \subfigure[]{
   \begin{minipage}[t]{0.5\linewidth}
   \centering
   \includegraphics[width=2.5in]{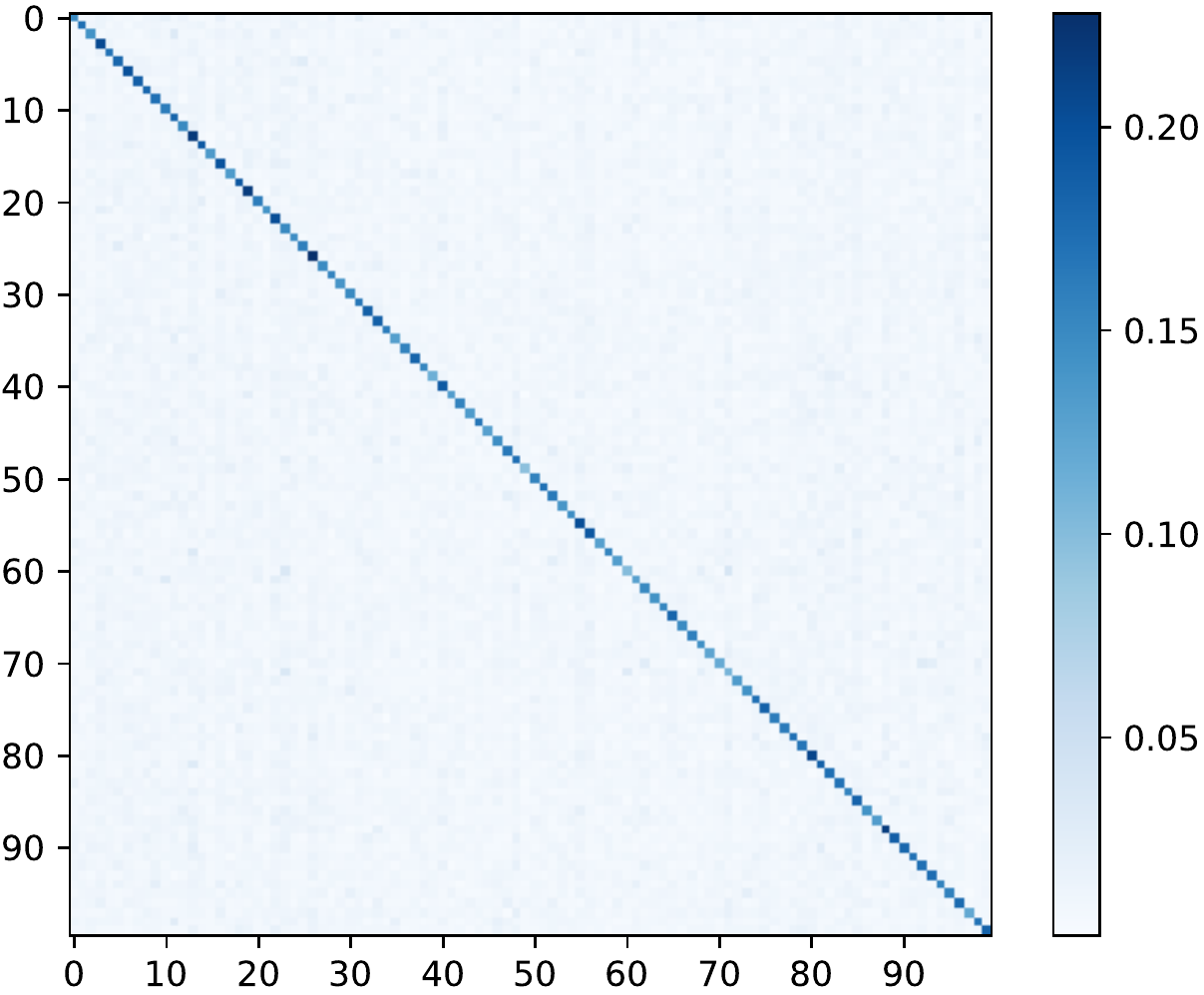}
   \end{minipage}%
   }%
   \centering
   \caption{Average of the Protective Label Distribution on CIFAR-10 (a) and CIFAR-100 (b).}
   \label{validate}
\end{figure}

\subsection{Proof of Theorem \ref{rho}}
\label{proofrho}
Generally, for epoch $t$, we denote $\Delta_{t}$ and $\gamma_{t}$ as the the unreliability degree and ambiguity degree of the the soft labels of the unlabeled data.
The model will achieve the accuracy of epoch $t$, $\rho(\Delta_{t},\gamma_{t})$.
We refer to it as $\rho_{t}$ for simplicity.
Then, for next epoch $t+1$,  the unreliability degree and ambiguity degree of next epoch can be estimated $\Delta_{t+1} \leq 1-\rho_{t}$.
The estimation of ambiguity degree is a little bit more complicated.
Incorrect label share equal probability $\gamma_{t+1}$ beacuse we assume the noise is uniformly distributed.
Then we have 
\begin{equation*}
   \Delta_{t+1}+(c-1)\gamma_{t+1}=k.
\end{equation*}
Further,
\begin{equation*}
   \gamma_{t+1} \leq \frac{c-k-\rho_{t}}{c-1}.
\end{equation*}
For $\rho(\Delta,\gamma)$ monotonically decreases, with the upper bounds on $\Delta_{t+1}$ and $\gamma_{t+1}$, we can get a lower bound on $\rho_{t+1}$ as 
\begin{equation*}
   \begin{aligned}
   \rho_{t+1} &= \rho(\Delta_{t+1},\gamma_{t+1})
   \\
   & \geq \rho(1-\rho_{t},\frac{c-k-\rho_{t}}{c-1}).
   \end{aligned}
\end{equation*}
If $\rho_{\text{final}} = \lim_{t \to \infty} \rho_{t}$ exists, it must satisfy the fix point equation,
\begin{equation*}
   x=\rho(1-x,\frac{c-k-x}{c-1}).
\end{equation*}

Next, we prove that if $\rho(\Delta,\gamma)$ is $k_{L}$-\textit{Lipschitz} continuous ($k_{L}<1-\frac{1}{c}$), then $\rho_{\text{final}}$ exists and is unique.
We define 
\begin{equation*}
   \psi: (\Delta,\gamma) \to (1-\rho(\Delta,\gamma),\frac{c-k}{c-1}-\frac{\rho(\Delta,\gamma)}{c-1}),
\end{equation*}
where $(\Delta, \gamma) \in [0,1]^{2}$.
$l_1$-norm is employed as the norm on $[0,1]^{2}$ and denote $d(\cdot,\cdot)$ as the distance function.
We want to show that $\psi$ is a \textit{contractive mapping}.
For $(\Delta_1,\gamma_1), (\Delta_2,\gamma_2) \in [0,1]^{2}$,
\begin{equation*}
   \begin{aligned}
      &d(\psi(\Delta_1,\gamma_1), \psi(\Delta_2,\gamma_2))\\
   =&(1+\frac{1}{c-1})|\rho(\Delta_1,\gamma_1)-\rho(\Delta_2,\gamma_2)| \\
   \leq & (1+\frac{1}{c-1}) \cdot k_{L} d((\Delta_1,\gamma_1),(\Delta_2,\gamma_2))
   \end{aligned} 
\end{equation*}
where $(1+\frac{1}{c-1}) \cdot k_{L} \in [0,1)$.
So $\psi$ is a \textit{contractive mapping} and there is a unique fixed point $(\Delta, \gamma)$ that $\psi((\Delta, \gamma))=(\Delta, \gamma)$.
That means $\rho_{\text{final}}$ exists and is unique.


\subsection{Overall distribution of the Customized Soft Labels}
\label{validate_2}
As illustrated in Figure \ref{validate}, the horizontal axis represents the ground-truth label while the vertical axis represents the mean of the soft labels.
The diagonal can be seen as the degree of correctly predicted labels.
We can see that the ground-truth label is dominant in the soft labels.
On the other side, the figure can be seen as a simple measure of the similarity between labels.

\subsection{Hyperparameters in (\ref{final})}
\label{hyper}
We set the number of the random labels \ref{srnd} as 3 while $k$ in $\Delta$ and $\gamma$ as 4 in all experiments.
The punishment factor $\alpha_1$ ranges from 0 to 0.4 and the compensation factor $\alpha_2$ ranges from 0.9 to 1.3.
The weight of the random labels $\alpha_3$ is from 1.6 to 2.3.
In detail, we observe the following phenomenons:
\begin{itemize}
   \item As mentioned in \ref{introduction}, we cannot take accuracy as the sole criterion to evaluate the teacher model. $\gamma$ defined in (\ref{gamma}) can be seen as the coarse measure of the imbalance of the label. For the student model, the smaller $\gamma$ could mean the better performance if the number of random labels is fixed.
   \item Both the punishment and the random labels are applied to decrease the top-$1$ accuracy within a reasonable range. For the simple dataset, large punishment are needed to protect privacy and for the complicated dataset like CIFAR-100, we enhance the ratio of random labels to reduce the effect of similar labels.
   \item The customized soft labels are very different from the ground-truth labels, which means low accuracy. In the customized soft, the average degree of the ground-truth labels is around 0.2. In common sense, such soft labels are considered ineffective.
\end{itemize}

\begin{table}[t]
   \caption{Classification accuracies for PLL.}
   \label{PLL}
   \begin{center}
   \begin{small}
\begin{tabular}{c|cc|c}
   \hline Dataset & $\Delta$ & $\gamma$ & Student \\
   \hline \multirow{9}{*}{ CIFAR-10 } & $0.1$ & $0.1$ & $\mathbf{93.98}$ \\
   & $0.1$ & $0.3$ & $\mathbf{93.38}$ \\
   & $0.1$ & $0.5$ & $\mathbf{91.94}$ \\
   \cline { 2 - 4 } & $0.3$ & $0.1$ & $\mathbf{90.38}$ \\
   & $0.3$ & $0.3$ & $\mathbf{88.59}$ \\
   & $0.3$ & $0.5$ & $\mathbf{86.28}$ \\
   \cline { 2 -4 } & $0.5$ & $0.1$ & $\mathbf{85.11}$ \\
   & $0.5$ & $0.3$ & $\mathbf{82.05}$ \\
   & $0.5$ & $0.5$ & $\mathbf{77.95}$ \\
   \hline \hline \multirow{9}{*}{ CIFAR-100 } & $0.01$ & $0.1$ & $\mathbf{74.19}$ \\
   & $0.1$ & $0.01$ & $\mathbf{73.16}$ \\
   & $0.1$ & $0.01$ & $\mathbf{72.29}$ \\
   \cline { 2 - 4 } & $0.05$ & $0.1$ & $\mathbf{68.08}$ \\
   & $0.3$ & $0.05$ & $\mathbf{66.28}$ \\
   & $0.3$ & $0.05$ & $\mathbf{63.69}$ \\
   \cline { 2 - 4 } & $0.1$ & $0.1$ & $\mathbf{61.18}$ \\
   & $0.5$ & $0.1$ & $\mathbf{58.6}$ \\
   & $0.5$ & $0.1$ & $\mathbf{52.41}$ \\
   \hline \hline
\end{tabular}
   \end{small}
   \end{center}
   \vskip -0.1in
\end{table}

\begin{figure}
   \centering 
   \includegraphics[width=2.95in]{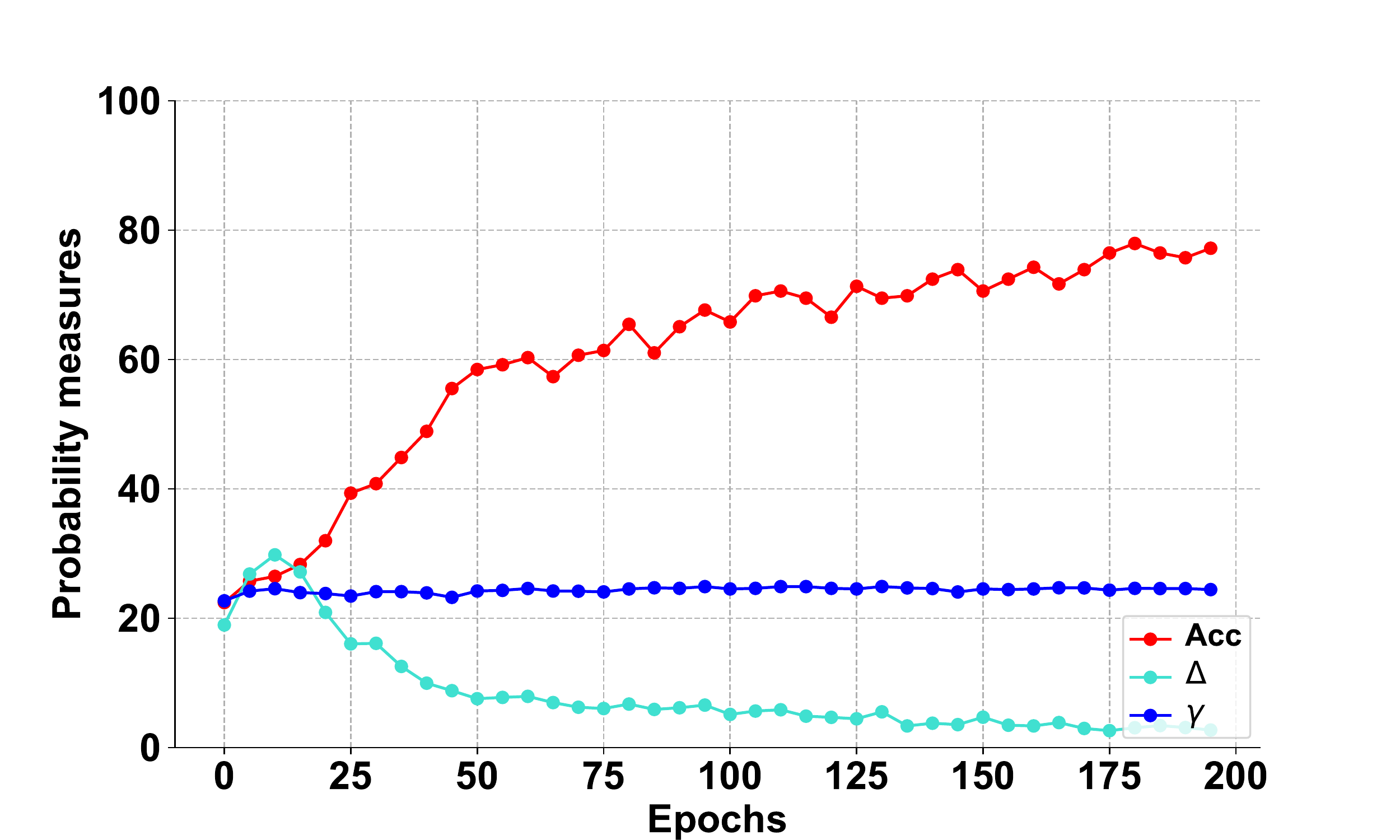}
   \caption{The curve of the indicators in the alternate training.} 
   \label{resun}
\end{figure}

\subsection{Experiments of leaerning with unlabeled data}
\label{unlabeled}
The predictive model label the unlabeled data and then learn with all data alternately.
As shown in Figure \ref{resun}, Acc (the accuracy of the model) improves as the soft labels envolve.
Accordingly, $\Delta$ decreases and $\gamma$ remain unchanged, which means the soft labels of the model are more effective.
The figure shows the dynamics in the training with unlabeled data, which is consistent with the theory in \ref{4.3}.

\subsection{Details in the experiments of weakly-supervised learning}
\label{dwsl}
In all experiments, the model architecture is WideResNet28$\times$2 architecture.
But there are some notations for weakly-supervised learning.
For PLL, it is a common strategy to zeroize the soft labels that are not in the candicate set.
For leaerning with unlabeled data, we wram-up the model for 5 epochs.
Then, we train with all data every epoch and label the unlabeled data with the model every 5 epochs.
CIFAR-10 are divided into labeled set, unlabeled set, testing set in
the ratio of 1:19:4.

\end{document}